\documentclass[pdflatex,iicol]{sn-jnl}
\usepackage{changes}
\usepackage{graphicx}%
\usepackage{multirow}%
\usepackage{amsmath,amssymb,amsfonts}%
\usepackage{amsthm}%
\usepackage{mathrsfs}%
\usepackage[title]{appendix}%
\usepackage{xcolor}%
\usepackage{textcomp}%
\usepackage{manyfoot}%
\usepackage{booktabs}%
\usepackage{algorithm}%
\usepackage{algorithmicx}%
\usepackage{algpseudocode}%
\usepackage{listings}%
\usepackage[sort]{natbib}%
\usepackage{pifont}
\newcommand{\cmark}{\textcolor{green!60!black}{\ding{51}}}  
\newcommand{\xmark}{\textcolor{red!70!black}{\ding{55}}}    
\usepackage{adjustbox}
\usepackage{tikz}
\usepackage[edges]{forest}
\usetikzlibrary{mindmap}
\usetikzlibrary{trees}
\definecolor{hidden-draw}{RGB}{20,68,106}
\definecolor{hidden-pink}{RGB}{255,245,247}
\usepackage{makecell}
\usepackage{array}
\usepackage[caption=false,font=normalsize,labelfont=sf,textfont=sf]{subfig}
\usepackage{tabularx}   
\usepackage{stfloats}
\usepackage{url}
\usepackage{verbatim}
\usepackage{hyperref}

\tikzstyle{my-box}=[
    rectangle,
    draw=hidden-draw,
    rounded corners,
    text opacity=1,
    minimum height=1.5em,
    minimum width=5em,
    inner sep=2pt,
    align=left,
    fill opacity=.5,
    line width=0.8pt,
]
\tikzstyle{level1}=[my-box, minimum height=1.5em,
    fill=gray!30, text=black, align=left,font=\normalsize,
    inner xsep=2pt,
    inner ysep=4pt,
    line width=0.8pt,
]
\tikzstyle{level2}=[my-box, minimum height=1.5em,
    fill=brown!10!white!90, text=black, align=left,font=\normalsize,
    inner xsep=2pt,
    inner ysep=4pt,
    line width=0.8pt,
]
\tikzstyle{level3}=[my-box, minimum height=1.5em,
    fill=brown!5!yellow!5!white!90, text=black, align=left,font=\normalsize,
    inner xsep=2pt,
    inner ysep=4pt,
    line width=0.8pt,
]
\tikzstyle{level4}=[my-box, minimum height=1.5em,
    fill=brown!10!yellow!10!white!80, text=black, align=left,font=\normalsize,
    inner xsep=2pt,
    inner ysep=4pt,
    line width=0.8pt,
]
\tikzstyle{leaf}=[my-box, minimum height=1.5em,
    fill=green!18, text=black, align=left,font=\normalsize,
    inner xsep=2pt,
    inner ysep=4pt,
    line width=0.8pt,
]

\tikzstyle{leaf1}=[my-box, minimum height=1.5em,
    fill=orange!18, text=black, align=left,font=\normalsize,
    inner xsep=2pt,
    inner ysep=4pt,
    line width=0.8pt,
]

\tikzstyle{leaf2}=[my-box, minimum height=1.5em,
    fill=yellow!30, text=black, align=left,font=\normalsize,
    inner xsep=2pt,
    inner ysep=4pt,
    line width=0.8pt,
]

\theoremstyle{thmstyleone}%
\theoremstyle{thmstyletwo}%

\theoremstyle{thmstylethree}%

\begin{document}

\title[Article Title]{A Survey of Multimodal Hallucination Evaluation and Detection}


\author[1,2]{\fnm{Zhiyuan} \sur{Chen}}\email{chenzhiyuan21@mail.ucas.ac.cn}

\author[1,2]{\fnm{Yuecong} \sur{Min}}\email{minyuecong@ict.ac.cn}

\author[1,2]{\fnm{Jie} \sur{Zhang}}\email{zhangjie@ict.ac.cn}

\author[1,2]{\fnm{Bei} \sur{Yan}}\email{yanbei23s@ict.ac.cn}

\author[3]{\fnm{Jiahao} \sur{Wang}}\email{wangjiahao50@huawei.com}

\author[3]{\fnm{Xiaozhen} \sur{Wang}}\email{jasmine.xwang@huawei.com}

\author[1,2]{\fnm{Shiguang} \sur{Shan}}\email{sgshan@ict.ac.cn}

\affil[1]{\orgdiv{State Key Laboratory of AI Safety}, \orgname{Institute of Computing Technology, \\ Chinese Academy of Sciences (CAS)}, \orgaddress{\city{Beijing}, \postcode{100190}, \country{China}}}

\affil[2]{\orgname{University of Chinese Academy of Sciences}, \orgaddress{\city{Beijing}, \postcode{100049}, \country{China}}}

\affil[3]{\orgdiv{Trustworthy Technology and Engineering Laboratory}, \orgname{Huawei}, \orgaddress{\city{Shenzhen}, \country{China}}}


\abstract{Multi-modal Large Language Models (MLLMs) have emerged as a powerful paradigm for integrating visual and textual information, supporting a wide range of multi-modal tasks. However, these models often suffer from hallucination, producing content that appears plausible but contradicts the input content or established world knowledge. This survey offers an in-depth review of hallucination evaluation benchmarks and detection methods across Image-to-Text (I2T) and Text-to-image (T2I) generation tasks. Specifically, we first propose a taxonomy of hallucination based on faithfulness and factuality, incorporating the common types of hallucinations observed in practice. Then we provide an overview of existing hallucination evaluation benchmarks for both T2I and I2T tasks, highlighting their construction process, evaluation objectives, and employed metrics. Furthermore, we summarize recent advances in hallucination detection methods, which aims to identify hallucinated content at the instance level and serve as a practical complement of benchmark-based evaluation. Finally, we highlight key limitations in current benchmarks and detection methods, and outline potential directions for future research.}

\keywords{Multi-modal large language model, Vision-language model, Diffusion model, Hallucination evaluation, Hallucination detection}

\maketitle

\section{Introduction}\label{sec:intro}

Over the past few years, Multi-Modal Large Language Models (MLLMs) have demonstrated remarkable advancements in bridging visual and textual data, supporting a wide range of multi-modal understanding and generation tasks. Image-to-Text (I2T) models such as GPT-4o~\citep{openai2024hello-gpt4o}, Gemini~\citep{team2023gemini}, and Qwen-VL~\citep{wang2024qwen2} excel at tasks like visual question answering and image captioning, achieving robust image recognition and reasoning capabilities without relying on external tools. Conversely, Text-to-Image (T2I) models like Stable Diffusion~\citep{rombach2022high} and DALL-E~\citep{ramesh2021zero} have made strides in generating high-quality images that align with user-specified content or artistic styles through textual prompts. Despite recent progress, both I2T and T2I models continue to face several significant challenges, such as robustness to distribution shifts~\citep{zhang2024b,guo2024efficient,aafaq2022language}, susceptibility to attacks~\citep{zhang2025modality,fan2024stealthy}, and hallucination~\citep{huang2023t2i,bai2024hallucination,lan2024survey,liu2024survey}. In particular, hallucination arises when models generate outputs that seem plausible but deviate from the given input or factual knowledge.

The hallucination problem remains a significant challenge to the development of large models. Early studies primarily focus on hallucination within I2T models, emphasizing inconsistencies between generated textual outputs and their corresponding visual inputs~\citep{rohrbach2018object,Li2023EvaluatingOH}. More recent works have extended this focus to T2I models~\citep{Chen2024UnifiedHD,Hu2023TIFAAA}, where hallucinations manifest as misalignments between generated visual outputs and textual prompts. This survey focuses on the hallucinations across both I2T and T2I paradigms of MLLMs, categorizing them into faithfulness hallucinations and factuality hallucinations. Faithfulness hallucination involves inconsistencies between model outputs and user inputs or prior output, such as misidentified objects or images that do not correspond to textual prompts. In contrast, factuality hallucination refers to contradictions between model outputs and established world knowledge, exemplified by incorrect landmark recognition or violations of medical plausibility in X-ray imaging tasks.

\begin{figure}[!t]
\centering
\includegraphics[width=\columnwidth]{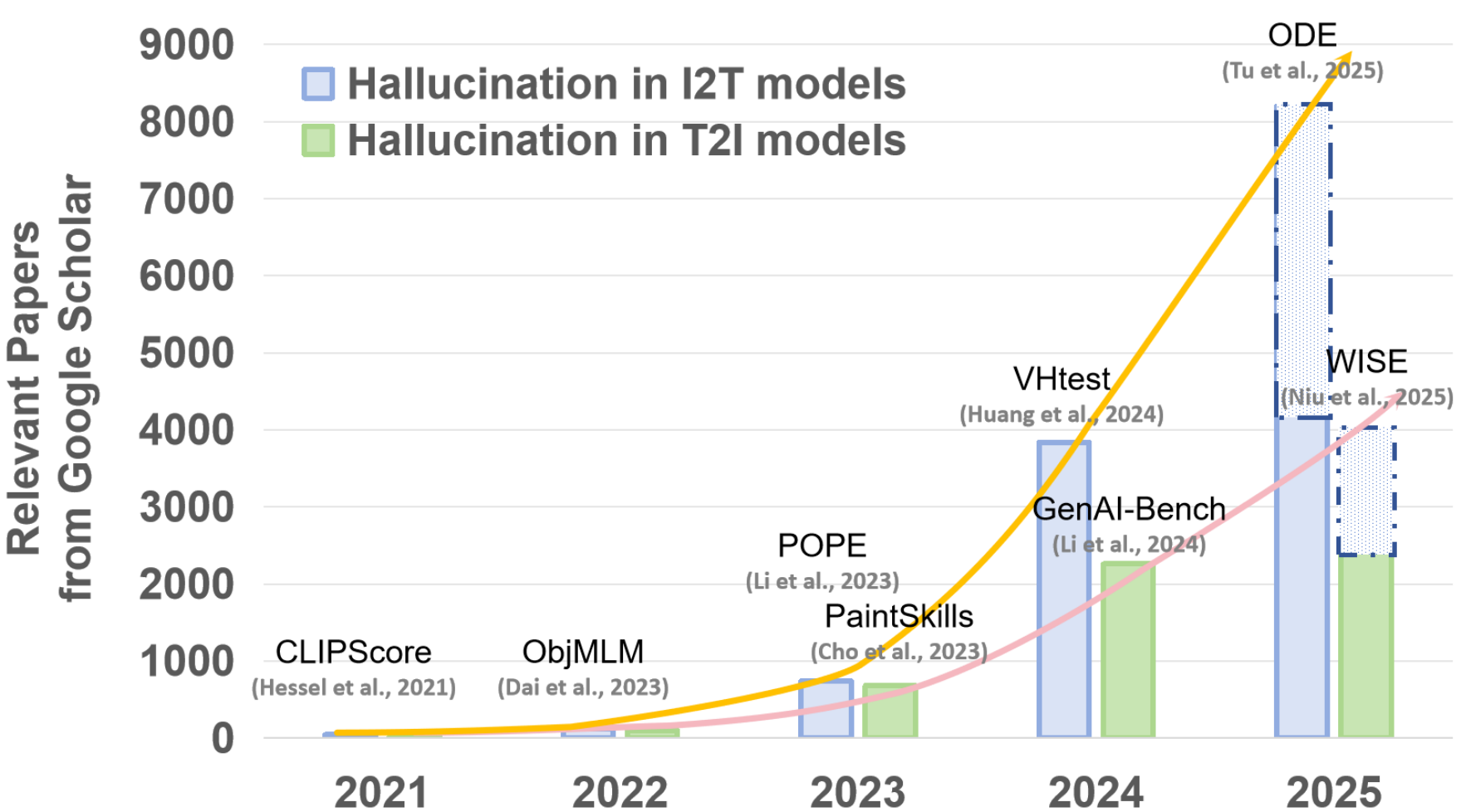}

\caption{Trends in the number of relevant papers on I2T and T2I hallucination evaluation from Google Scholar, highlighting the rapid growth in recent years. Dashed lines indicate approximate predictions.}
\label{tendency}
\end{figure}

As shown in Fig.~\ref{tendency}, hallucination evaluation and benchmark construction have received increasing attention, resulting in numerous benchmarks for assessing hallucinations in both I2T models~\citep{Li2023EvaluatingOH, Wu2024EvaluatingAA, fu2023mme, seth2024hallucinogen, zhou2023don} and T2I models~\citep{feng2023training, gokhale2022benchmarking, huang2024t2i, Hu2023TIFAAA} from various perspectives. This paper systematically reviews existing I2T and T2I benchmarks targeting faithfulness and factuality hallucinations. Specifically, we provide a detailed comparison across benchmarks in terms of data sources, evaluation tasks, image-text pair construction, and hallucination types, highlighting the emerging trends towards automatic construction and fine-grained evaluation. Furthermore, we give a unified perspective on hallucination evaluation across T2I and I2T benchmarks, summarizing their commonalities as well as differences.\par
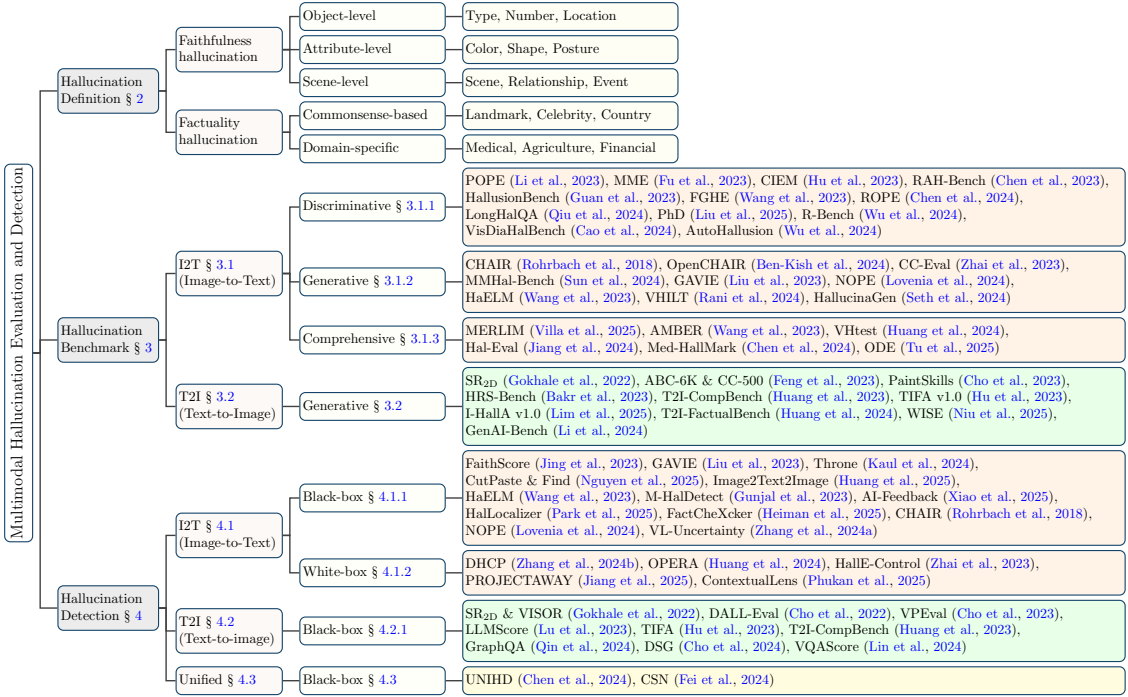
\begin{figure*}[t]
    \centering
    \resizebox{0.93\textwidth}{!}{
        \begin{forest}
            forked edges,
            for tree={
                grow=east,
                reversed=true,
                anchor=base west,
                parent anchor=east,
                child anchor=west,
                base=left,
                font=\large,
                rectangle,
                draw=hidden-draw,
                rounded corners,
                align=left,
                minimum width=4em,
                edge+={darkgray, line width=1pt},
                s sep=3pt,
                inner xsep=2pt,
                inner ysep=3pt,
                line width=0.8pt,
                ver/.style={rotate=90, child anchor=north, parent anchor=south, anchor=center},
            },
            where level=1{text width=6.8em,font=\normalsize,}{},
            where level=2{text width=7.2em,font=\normalsize,}{},
            where level=3{text width=10em,font=\normalsize,}{},
            where level=4{text width=5em,font=\normalsize,}{},
            [
                Multimodal Hallucination Evaluation and Detection, ver
                [
                    Definition
                    \S~\ref{sec:alldef}, level1
                    [
                        Model \\Definition \S~\ref{sec:modeldef},level2
                        [
                            Image-to-Text \\(I2T) Models \S~\ref{sec:i2tmodel},level3
                            [Visual Encoder{, }Vision-language Adapters{, }Text Decoder{, }Training~\citep{danish2025comprehensive}, level4,text width=40em]
                        ]
                        [
                            Text-to-Image \\(T2I) Models \S~\ref{sec:t2imodel}, level3
                            [Text Encoder{, }Latent Encoder{, }Backbone{, }Training~\citep{shenefficient,ma2025efficient}, level4,text width=40em]
                        ]
                    ]
                    [
                        Hallucination \\Definition \S~\ref{sec:def},level2
                        [
                            Faithfulness \\Hallucination,level3
                            [
                                Object-level, level3, text width=10em
                                [
                                    Type~\citep{rohrbach2018object,Hu2023TIFAAA}{, }Number~\citep{chen2024multi}{, }\\Location~\citep{seth2024hallucinogen}, leaf1, text width=35em
                                ]
                            ]
                            [
                                Attribute-level, level3, text width=10em
                                [Color~\citep{fu2023mme,feng2023training}{, }Shape~\citep{huang2024visual,huang2023t2i}{, }\\Posture~\citep{yan2025shale}, leaf1, text width=35em]
                            ]
                            [
                                Scene-level, level3, text width=10em
                                [Scene~\citep{park2025halloc,bakr2023hrs}{, }Relationship~\citep{Wu2024EvaluatingAA}{, }\\Event~\citep{Jiang2024HalEvalAU}, leaf1, text width=35em]    
                            ] 
                        ]
                        [
                            Factuality    \\Hallucination, level3
                            [
                                Commonsense-based, level3, text width=10em
                                [Landmark~\citep{yan2025shale}{, }Celebrity~\citep{fu2023mme}{, }\\Daily Objects~\citep{huang2024t2i}, leaf1, text width=35em]
                            ]                
                            [
                                Domain-specific, level3, text width=10em
                                [Medical~\citep{ayaz2024medvlm}{, }Agriculture~\citep{arshad2025leveraging}{, }\\Scientific Facts~\citep{lim2025evaluating}, leaf1, text width=35em]
                            ]         
                        ]            
                    ]            
                ]     
                [                            
                    Hallucination\\ Benchmark
                    \S~\ref{sec:benchmark}, level1
                    [   
                        I2T  \S~\ref{beni2t}\\(Image-to-Text), level2
                        [
                            Discriminative \S~\ref{i2tdis}, level3
                            [
                                POPE~\citep{Li2023EvaluatingOH}{, }MME~\citep{fu2023mme}{, }CIEM~\citep{hu2023ciem}{, }RAH-Bench~\citep{chen2023mitigating}{, }\\HallusionBench~\citep{Guan2023HallusionBenchAA}{, }FGHE~\citep{Wang2023MitigatingFH}{, }ROPE~\citep{chen2024multi}{, }\\LongHalQA~\citep{qiu2024longhalqa}{, }PhD~\citep{liu2025phd}{, }R-Bench~\citep{Wu2024EvaluatingAA}{, }\\VisDiaHalBench~\citep{cao2024visdiahalbench}{, }AutoHallusion~\citep{wu2024autohallusion}, leaf1, text width=47em
                            ]
                        ]
                        [
                            Generative \S~\ref{i2tgen}, level3
                            [
                                CHAIR~\citep{rohrbach2018object}{, }OpenCHAIR~\citep{ben2024mitigating}{, }CC-Eval~\citep{zhai2023halle}{, }\\MMHal-Bench~\citep{sun2024aligning}{, }GAVIE~\citep{Liu2023MitigatingHI}{, }NOPE~\citep{lovenia2024negative}{, }\\HaELM~\citep{wang2023evaluation}{, }VHILT~\citep{rani2024visual}{, }HallucinaGen~\citep{seth2024hallucinogen}, leaf1, text width=47em
                            ]
                        ]
                        [
                            Comprehensive \S~\ref{i2tcom}, level3
                            [
                                MERLIM~\citep{villa2025behind}{, }AMBER~\citep{wang2023amber}{, }VHtest~\citep{huang2024visual}{, }\\Hal-Eval~\citep{Jiang2024HalEvalAU}{, }Med-HallMark~\citep{chen2024detecting}{, }ODE~\citep{tu2025ode}, leaf1, text width=47em
                            ]
                        ]   
                    ]
                    [
                        T2I   \S~\ref{bent2i}\\(Text-to-Image), level2
                        [
                            Generative \S~\ref{bent2i}, level3
                            [
                                $\text{SR}_{\text{2D}}$~\citep{gokhale2022benchmarking}{, }ABC-6K \& CC-500~\citep{feng2023training}{, }PaintSkills~\citep{cho2023dall}{, }\\HRS-Bench~\citep{bakr2023hrs}{, }T2I-CompBench~\citep{huang2023t2i}{, }TIFA v1.0~\citep{Hu2023TIFAAA}{, }\\I-HallA v1.0~\citep{lim2025evaluating}{, }T2I-FactualBench~\citep{huang2024t2i}{, }WISE~\citep{niu2025wise}{, }\\GenAI-Bench~\citep{li2024evaluating}, leaf, text width=47em
                            ]
                        ]
                    ]
                ]         
                [
                    Hallucination\\Detection \S~\ref{sec:det}, level1
                    [
                        I2T   \S~\ref{deti2t}\\(Image-to-Text), level2
                        [
                            Black-box \S~\ref{i2tbla}, level3
                            [
                                FaithScore~\citep{Jing2023FaithScoreFE}{, }GAVIE~\citep{Liu2023MitigatingHI}{, }Throne~\citep{kaul2024throne}{, }\\CutPaste \& Find~\citep{nguyen2025cutpaste}{, }Image2Text2Image~\citep{huang2025image2text2image}{, }\\HaELM~\citep{wang2023evaluation}{, }M-HalDetect~\citep{Gunjal2023DetectingAP}{, }AI-Feedback~\citep{xiao2025detecting}{, }\\HalLocalizer~\citep{park2025halloc}{, }FactCheXcker~\citep{heiman2025factchexcker}{, }CHAIR~\citep{rohrbach2018object}{, }\\NOPE~\citep{lovenia2024negative}{, }VL-Uncertainty~\citep{zhang2024vl}, leaf1, text width=47em
                            ]
                        ]
                        [
                            White-box \S~\ref{i2twhi}, level3
                            [
                                DHCP~\citep{zhang2024dhcp}{, }OPERA~\citep{huang2024opera}{, }HallE-Control~\citep{zhai2023halle}{, }\\PROJECTAWAY~\citep{jianginterpreting}{, }ContextualLens~\citep{phukan2025beyond}, leaf1, text width=47em
                            ]
                        ]
                    ]
                    [
                        T2I   \S~\ref{dett2i}\\(Text-to-image), level2
                        [
                            Black-box \S~\ref{t2ibla}, level3
                            [
                                $\text{SR}_{\text{2D}}$ \& VISOR~\citep{gokhale2022benchmarking}{, }DALL-Eval~\citep{Cho2022DALLEVALPT}{, }VPEval~\citep{cho2023visual}{, }\\LLMScore~\citep{lu2023llmscore}{, }TIFA~\citep{Hu2023TIFAAA}{, }T2I-CompBench~\citep{huang2023t2i}{, }\\GraphQA~\citep{qin2024evaluating}{, }DSG~\citep{cho2024davidsonian}{, }VQAScore~\citep{lin2024evaluating}, leaf, text width=47em
                            ]
                        ]
                    ]
                    [
                        Unified  \S~\ref{detuni}, level2
                        [
                            Black-box \S~\ref{detuni}, level3
                            [
                                UNIHD~\citep{Chen2024UnifiedHD}{, }CSN~\citep{fei2024fine}, leaf2, text width=47em
                            ]
                        ]
                    ]
                ]
            ]
        \end{forest}
        }
    \centering

    \caption{Overview of the main structure and taxonomy presented in this survey.}
    \vspace{-5 pt}
    \label{taxonomy}
\end{figure*}

The flexible response formats of MLLMs present additional challenges for hallucination evaluation, particularly in free-form tasks such as image captioning and image generation. Therefore, hallucination detection methods play a crucial role in the construction of evaluation benchmarks by reducing reliance on costly human annotations and enabling scalable evaluation. Conversely, evaluation benchmarks provide annotated data and standardized metrics essential for the development and assessment of hallucination detection methods. To provide a more holistic perspective on the complementary relationship between evaluation and detection, we further provide a comprehensive summary of existing hallucination detection methods and discuss the feasibility of hallucination detection in I2T and T2I models from a unified perspective.\par

In a nutshell, this paper aims to advance hallucination evaluation in MLLMs through a comprehensive survey of benchmarks and detection methods for both I2T and T2I tasks. In contrast to existing surveys~\citep{huang2023t2i,bai2024hallucination,lan2024survey,liu2024survey} that primarily focus on hallucinations arising in I2T tasks, we broaden the scope to include hallucinations in T2I tasks and provide a unified overview of both faithfulness and factuality hallucinations. Furthermore, we offer a systematic summary of existing hallucination evaluation and detection methods and introduce a clear categorization framework. For example, we categorize detection methods into black-box and white-box approaches with different granularity. We also highlight recent efforts towards unified hallucination evaluation and detection across both I2T and T2I models, contributing to a more comprehensive understanding of hallucination phenomena.

\textbf{Organization of this survey.}  
To facilitate a clearer understanding of existing research, we organize this survey as follows. As shown in Fig.~\ref{taxonomy}, we begin formally defining the models and the concept of hallucinations within this context in Sec.~\ref{sec:alldef}. Next, we analyze and summarize existing hallucination benchmarks for I2T and T2I models in Sec.~\ref{sec:benchmark}, focusing on aspects such as hallucination types, construction methodologies, and evaluation metrics, while highlighting recent trends in benchmark development. Building on this foundation, we categorize existing MLLMs hallucination detection methods and investigate the feasibility of unified detection approaches in Sec.~\ref{sec:det}. Finally, we discuss current limitations and challenges in hallucination evaluation, and suggest promising directions for future research.\par

\section{Definition}
\label{sec:alldef}
In this chapter, we provide the conceptual basis by defining multimodal models and hallucination.
Since hallucinations arise from complex interactions between model architecture, training dynamics, and modal alignment, 
we first provide precise definitions of both the I2T and T2I models and outline the components and mechanisms that may guide their behavior in Sec.~\ref{sec:modeldef}.
Sec.~\ref{sec:def} then formalizes the definition of hallucination in MLLMs, categorizing them into faithfulness and factuality hallucinations.

\subsection{Definition of Models}
\label{sec:modeldef}

This section introduces the two primary categories of multimodal generative systems examined in this survey: 
I2T models and T2I models. Sec.~\ref{sec:i2tmodel} and Sec.~\ref{sec:t2imodel} describe I2T and T2I models respectively, 
focusing on their architectures, multimodal interaction mechanisms and training. These definitions support the later
discussion on hallucination causes.

\subsubsection{Image-to-Text (I2T) Models}
\label{sec:i2tmodel}
Image-to-text (I2T) models have transitioned from specialized, task-specific architectures to unified, general-purpose systems capable of complex reasoning, 
detailed captioning, and visual question answering (VQA). Representative models such as LLaVA~\citep{liu2023visual}, Qwen-VL~\citep{wang2024qwen2}, and GPT-4V~\citep{openai_gpt4v_2023} typically comprise a visual encoder, an alignment adapter, and an LLM decoder.
Architectural constraints in these components, such as resolution limits, are distinct causal factors for hallucination. Consequently, errors arise from both inherited LLM tendencies, like
over-reliance on language priors~\citep{huang2024opera},  and unique mismatches between visual representations and the decoder's semantic space.

\textit{Visual Encoder.}
The evolution of visual encoders in I2T models reflects a shift from static, low-resolution global alignment toward dynamic, high-fidelity feature extraction 
intended to reduce sensory hallucination. Early contrastive architectures such as CLIP~\citep{Radford2021LearningTV} treat the image as a ``bag of features,'' identifying objects without enforcing spatial structure. 
Their fixed low-resolution inputs also act as low-pass filters that remove fine details and trigger object- or attribute-level hallucinations. Recent works such as LLaVA-NeXT~\citep{liu2024llava} and Qwen2-VL~\citep{wang2024qwen2} 
adopt dynamic-resolution strategies (\emph{e.g.}, AnyRes grid decomposition), which preserve high-frequency details by processing images at native scales. However, these solutions introduce new trade-offs: grid decomposition may break object continuity and create duplicate predictions, 
while variable-length sequences can overwhelm the decoder with long token streams that suffer from information dilution~\citep{liu2024lost} or visual attention sinks~\citep{kangsee}.

\textit{Vision-language Adapter.}
Adapter design balances information preservation and token efficiency, directly influencing hallucination behaviors. MLP-based adapters such as LLaVA~\citep{liu2023visual} 
preserve visual detail but may induce faithfulness errors when modality gaps cause the model to rely on language priors~\citep{liang2022mind}. Q-former compression 
in BLIP-2~\citep{li2023blip} creates bottlenecks that can remove important cues and lead to non-existence hallucinations~\citep{yan2024vigor}. Deep fusion approaches such as 
Flamingo~\citep{alayrac2022flamingo} attempts closer interleaving of modalities, yet gating mechanisms may still allow the LLM's internal knowledge to dominate, resulting in
omission of visually grounded information~\citep{he2025cracking}.

\textit{Text Decoder.}
The text decoder often utilizes powerful decoder-only LLMs (\emph{e.g.}, Vicuna, LLaMA, Qwen) as the text generation backbone, a design choice that inextricably
links the model's linguistic capabilities to the root causes of hallucination~\citep{liu2024survey}. This reliance creates a conflict where strong
pre-trained language priors (\emph{e.g.}, ``fork'' following ``plate'') often override weaker visual signals, leading to object existence hallucinations~\citep{Li2023EvaluatingOH}.
This issue is exacerbated by ``sycophancy,'' a learned behavior from instruction tuning where models prioritize user intent over visual factuality, agreeing
with misleading premises (\emph{e.g.}, confirming the presence of a non-existent object) to remain ``helpful''~\citep{pi2025pointing}.
Furthermore, ``parametric knowledge interference'' occurs when the model's encyclopedic pre-training supersedes specific visual context, causing it to hallucinate
biography-style details or facts that are true in the world but false for the specific image at hand~\citep{zhu2024unraveling}.

\textit{Training.}
Training aligns visual and textual modalities, utilizing strategies ranging from single-stage adapter integration (\emph{e.g.}, LLaMA-Adapter~\citep{zhang2023llama}) to two-stage pipelines combining feature alignment with instruction tuning~\citep{liu2024improved,zhu2024minigpt}.
Hallucination is often ``trained into'' the model through biased data or misalignment of objectives. On the data format, an exclusive reliance on positive image-text pairs fosters a stubborn ``Yes-bias'', necessitating special designs in instruction tuning where models are explicitly taught to reject 
non-existence objects~\citep{Liu2023MitigatingHI}. Alignment is shifting from reinforcement learning from human feedback (RLHF~\cite{christiano2017deep}) to direct preference optimization (DPO~\cite{rafailov2023direct}) and variants like HA-DPO~\citep{zhao2023beyond}. By utilizing preference pairs that contrast accurate
versus hallucinatory outputs, HA-DPO directly penalizes ungrounded tokens to enforce visual factuality rather than mere conversational fluency.

\subsubsection{Text-to-Image (T2I) Models}
\label{sec:t2imodel}

Text-to-Image (T2I) models have evolved into high-capacity architectures capable of producing photorealistic and semantically aligned images
by mapping natural language to complex visual manifolds. Modern systems, such as Stable Diffusion 3~\citep{esser2024scaling} and DALL·E 3~\citep{betker2023improving}, 
leverage massive datasets and advanced alignment techniques to achieve high semantic fidelity. Architecturally, the field is shifting from U-Net backbones toward 
Diffusion Transformers (DiT)~\citep{peebles2023scalable}, comprising four core components: a text encoder for semantic extraction, a VAE for latent compression, a diffusion-based backbone 
for iterative denoising, and an alignment pipeline.

\textit{Text Encoder.} Text encoders convert prompts into structured embeddings capturing semantic and syntactic information. Early models relied on contrastive encoders like CLIP~\citep{Radford2021LearningTV}, 
whose global pooling often overlooks compositional constraints~\citep{koishigarina2025clip}. LLM-based encoders such as T5~\citep{chung2024scaling}, LLaMA~\citep{touvron2023llama}, and Baichuan~\citep{yang2023baichuan} 
provide richer syntactic awareness and better handle complex prompts, including negation, counting, and long-range dependencies~\citep{jiao2025detailmaster}. Dual-stream attention mechanisms, as in MMDiT~\citep{StableDiffusion3_2025}, 
enable cross-modal interaction while maintaining distinct representations, reducing faithfulness-related hallucinations.

\textit{Latent Encoder.} In latent diffusion models, a Variational Autoencoder (VAE) compresses high-dimensional images into a lower-dimensional latent space~\citep{rombach2022high}. The VAE outputs a latent distribution, sampled for reconstruction, with KL divergence regularizing it toward a standard Gaussian. Compression is lossy: fine textures and subtle details are discarded~\citep{relic2024lossy}, forcing the decoder to hallucinate missing information. Variants include standard VAEs and quantized VAEs (VQVAE~\citep{esser2021taming}, VQGAN~\citep{van2017neural}), sometimes augmented with adversarial losses.

\textit{Backbone.} The backbone predicts noise residuals or velocity fields that drive generation. While early U-Net architectures~\citep{ho2020denoising} suffer from structural hallucinations due to limited long-range interactions~\citep{lu2025towards},
the transition to Diffusion Transformers (DiT)~\citep{peebles2023scalable} mitigates this by employing global self-attention to maintain 
full-image receptive fields. This architecture scales effectively, as demonstrated by Flux.1~\citep{FLUX.1-dev_2025} and Stable Diffusion 3~\citep{esser2024scaling}.
Furthermore, state-of-the-art models increasingly adopt Rectified Flow~\citep{liuflow} to enforce straight generation trajectories, significantly 
improving stability and eliminating artifactual hallucinations during few-step synthesis.

\textit{Training.} Recent T2I training increasingly utilizes Rectified Flow Transformers~\citep{esser2024scaling,peebles2023scalable} rather than standard diffusion U-Nets,
optimizing a generative backbone to straighten noise-to-data trajectories via flow matching~\citep{liuflow,lipman2023flow}. The paradigm remains multi-phase: pre-training on massive
image-text pairs for coarse modality alignment, followed by high-resolution finetuning. However, human alignment now favors DPO over complex RLHF pipelines,
directly optimizing likelihoods against preference pairs without explicit reward modeling. For adaptation, parameter-efficient techniques like LoRA~\citep{hulora} and ControlNets~\citep{zhang2023adding} persist
but are now adapted for transformer architectures. Throughout these phases, stability is ensured via velocity-based constraints and classifier-free guidance augmentation,
enabling models to generate high-fidelity images that adhere strictly to complex semantic prompts.

\subsection{Definition of Hallucination}
\label{sec:def}
Hallucination is a general psychological term~\citep{bentall1990illusion} referring to a percept-like experience that occurs without an external stimulus, possesses the vividness of genuine perception, and is not under voluntary control~\citep{slade1988sensory}. In the context of artificial intelligence, hallucination typically refers to the generation of unfaithful or counterfactual content. This phenomenon has garnered significant attention across multiple fields, including natural language generation~\citep{ji2023survey}, computer vision~\citep{yu2018generative,rohrbach2018object}, and multi-modal modeling~\citep{bai2024hallucination}.\par
In the context of computer vision, the term ``hallucination'' originated in image super-resolution~\citep{baker2000hallucinating,liu2007face,wang2014comprehensive,huang2019wavelet} task and has since been applied to other image generation tasks, such as inpainting~\citep{yu2018generative,quan2024deep} and novel sample generation~\citep{hariharan2017low,yang2025image}. In practice, hallucination is leveraged to generate realistic images or to enhance generalization capabilities. However, unlike its beneficial role in generative tasks, hallucination in vision perception tasks is often undesirable. It typically refers to instances where the model produces false or misleading outputs that do not correspond to the input data. For example, in object detection~\citep{rohrbach2018object} or image caption~\citep{kayhan2021hallucination, piasco2021improving}, a model may hallucinate objects that are not present in the scene, leading to reduced reliability and potential risks in safety-critical applications such as autonomous driving~\citep{you2024v2x} or medical imaging~\citep{ayaz2024medvlm}.\par
With the rapid advancement of image generation techniques, the notion of hallucination has shifted to the alignment between generated content and the input prompt~\citep{aithal2024understanding,kim2024tackling,wang2024hallo3d}. Moreover, hallucination in I2T, \emph{i.e.}, visual perception~\citep{saleh2025building}, and T2I, \emph{i.e.}, visual generation~\citep{agnese2020survey}, tasks have become increasingly interrelated: generated images are used to evaluate the faithfulness of I2T models and vice versa. One interesting work attempts to detect hallucinations across both I2T and T2I tasks~\citep{Chen2024UnifiedHD} within a unified framework. To provide a more consistent viewpoint between T2I and I2T tasks, we adopt a unified definition of hallucination applicable to both I2T and T2I settings: \textit{any inconsistency between the generated content and either the input conditions or established world knowledge}. This definition is consistent with the psychological concept of hallucination as a perception without an external stimulus that contradicts reality and also provides a comprehensive viewpoint of recent works.

\begin{figure*}
    \centering
    \includegraphics[width=0.95\textwidth]{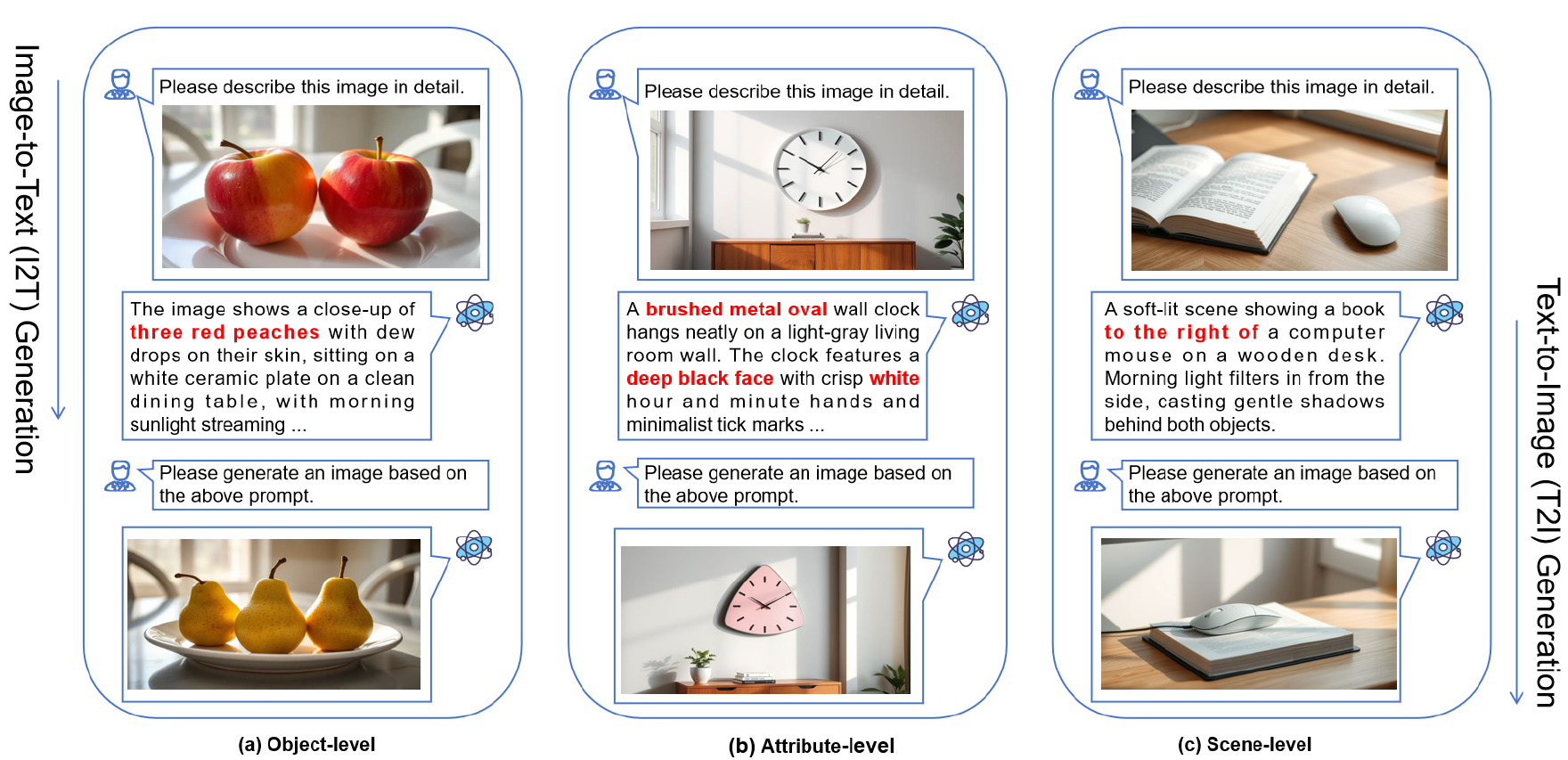}
    \caption{Examples of object-level (a), attribute-level (b), and scene-level (c) hallucinations (from left to right) in image-to-text (top) and text-to-image (bottom) tasks. Hallucinated responses are highlighted in red. All the images are generated by Stable Diffusion 1.5~\citep{rombach2022high}.}
    \label{faithfulness}
\end{figure*}

\subsubsection{Hallucinations in I2T Tasks}

\label{subsec:i2t}
\begin{figure*}
    \centering
    \includegraphics[width=0.95\textwidth]{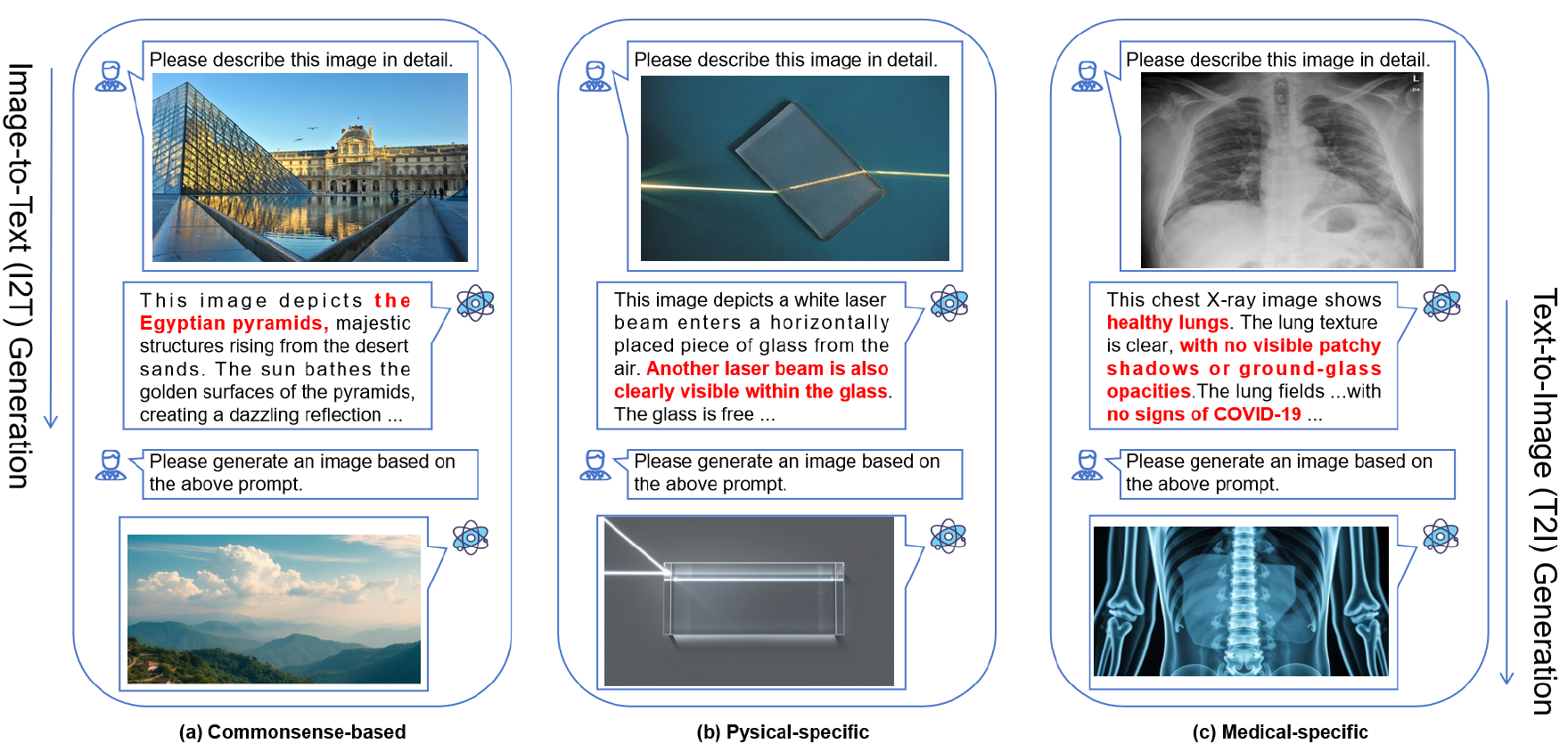}
    \vspace{-5pt}
    \caption{Examples of Commonsense-based (a), Physical-specific (b), and Medical-specific (c) hallucinations (from left to right) in image-to-text (top) and text-to-image (bottom) tasks. Hallucinated responses are highlighted in red. All the images are generated by Stable Diffusion 1.5~\citep{rombach2022high}.}
    \label{factuality}
\end{figure*}

Hallucination in Vision-Language Models (VLMs) refers to the generated text response being inconsistent with the input visual content or established world knowledge. Building on the taxonomy of hallucination in the context of LLM~\citep{huang2025survey}, we extend this framework to VLMs and categorize hallucinations into two types: faithfulness hallucination and factuality hallucination. The former captures inconsistencies between the generated contents and the visual inputs, while the latter refers to discrepancies between the generated contents and established world knowledge~\citep{fu2023mme,Guan2023HallusionBenchAA,seth2024hallucinogen}. These kinds of hallucinations may arise from factors like noisy training data, strong language priors, cross-modal misalignment, and other related issues~\citep{huang2025survey}. To better assess the reliability of recent VLMs, numerous studies have proposed hallucination benchmarks that evaluate different aspects of this phenomenon, including object hallucination~\citep{Li2023EvaluatingOH}, relationship hallucination~\citep{Wu2024EvaluatingAA}, attribute hallucination~\citep{fu2023mme} and so on. Fig.~\ref{faithfulness} and Fig.~\ref{factuality} present examples of different kinds of hallucinations in I2T tasks.

\textbf{Faithfulness Hallucinations.} Recent works on faithfulness hallucinations primarily focus on object hallucination~\citep{rohrbach2018object,bai2024hallucination,dai2023plausible,Li2023EvaluatingOH,ben2024mitigating,zhai2023halle,lovenia2024negative} and further categorize into category~\citep{chen2024multi,Wang2023MitigatingFH,liu2025phd,wang2023amber}, attribute~\citep{fu2023mme,chen2023mitigating,hu2023ciem,qiu2024longhalqa,wang2023amber}, and relationship~\citep{Wang2023MitigatingFH,chen2023mitigating,Wu2024EvaluatingAA,hu2023ciem,qiu2024longhalqa,villa2025behind,wang2023amber} hallucination. These types form a fine-to-fine-grained taxonomy. Category hallucination refers to errors in object classification. Attribute hallucination refers to the errors in attribute recognition of specific objects. Relationship hallucination requires recognizing object category and attributes, as well as objects' interrelations. However, hallucinations such as scene hallucination cannot be captured by this taxonomy. Inspired by the hierarchical structure of computer vision, ranging from image-level to pixel-level based on supervision strength~\citep{nesteruk2024survey}, we propose a taxonomy of faithfulness hallucination based on the granularity of visual content: object-level, attribute-level and scene-level. While
faithfulness hallucinations in LLMs manifesting as inconsistencies in instruction following, contextual grounding, or logical coherence~\citep{huang2025survey}, such granularity-based taxonomy reflects new perceptual challenges brought by multimodal.\par
\begin{itemize}
    \item \textit{Object-level} hallucinations involve the basic information of objects within the image, such as object type~\citep{rohrbach2018object}, object number~\citep{chen2024multi}, object localization~\citep{seth2024hallucinogen} and so on. For example, in Fig.~\ref{faithfulness}(a), the generated description (peaches) is inconsistent with the image content (apples).
    \item \textit{Attribute-level} hallucinations refer to the condition that the basic information of objects identified by VLMs is accurate, but the fine-grained intra-object content is wrong, which can be categorized into color~\citep{fu2023mme}, shape~\citep{huang2024visual} and so on. In Fig.~\ref{faithfulness}(b), the description of the clock is inconsistent with the image, particularly in terms of its material, shape, and color.
    \item \textit{Scene-level} hallucinations refer to the inconsistencies between global visual inputs and textual outputs. VLMs correctly identify the basic information of objects, but hallucinate in inter-object information such as object relationships~\citep{Wang2023MitigatingFH,chen2023mitigating,Wu2024EvaluatingAA}, object events~\citep{Jiang2024HalEvalAU}, scene~\citep{park2025halloc} and so on. In Fig.~\ref{faithfulness}(c), the spatial relation is described as ``to the right of'', which contradicts the actual positioning in the image (``to the left of'').
\end{itemize}

\textbf{Factuality Hallucinations.}
With the advancement of in-context learning and visual understanding capabilities of VLMs, they have been increasingly applied in practical scenarios including medical image analysis~\citep{ayaz2024medvlm}, autonomous driving~\citep{you2024v2x} and smart agriculture~\citep{arshad2025leveraging}. However, VLMs may produce hallucinations due to the insufficient grounding in established world knowledge~\citep{seth2024hallucinogen,chen2024detecting}. Existing benchmarks primarily focus on commonsense knowledge~\citep{fu2023mme,Guan2023HallusionBenchAA}, while some recent works concerning specific domain knowledge~\citep{chen2024detecting,seth2024hallucinogen} have emerged. Based on the source of established world knowledge, we categorized factuality hallucinations into commonsense-based and domain-specific hallucinations.
\begin{itemize}
    \item \textit{Commonsense-based} hallucinations refer to VLMs violating the general knowledge which is universally accepted by humans but usually implicitly stated, such as the appearance of landmarks~\citep{yan2025shale} or features of celebrities~\citep{fu2023mme}. For example, in Fig.~\ref{factuality}(a), VLM identifies the landmark in the image as the Egyptian pyramids, while it is actually the Louvre Museum in France.
    \item \textit{Domain-specific} hallucinations refer to inconsistencies between the outputs of VLMs and factual information when processing images from specific domains. In Fig.~\ref{factuality}(b), the VLM fails to recognize the refraction of the laser beam within the glass and instead hallucinates an additional laser beam inside the glass. In Fig.~\ref{factuality}(c), the VLM identifies the lungs affected by COVID-19 in the X-ray image as healthy lungs.
\end{itemize}

\subsubsection{Hallucinations in T2I Tasks}
\label{subsec:t2i}
Hallucinations in T2I tasks refer to the generated image content being inconsistent with the input text prompt or established world knowledge. By summarizing existing T2I hallucination benchmarks~\citep{gokhale2022benchmarking,feng2023training,saharia2022photorealistic,cho2023dall,bakr2023hrs,Chen2024UnifiedHD,fei2024fine}, we categorize hallucinations in T2I tasks into faithfulness hallucination and factuality hallucination, similar to the VLMs. Faithfulness hallucination refers to discrepancies between generated image content and input text prompt, while factuality hallucination refers to discrepancies between established world knowledge and input text prompt. Numerous studies have proposed benchmarks to evaluate faithfulness hallucination in T2I models~\citep{gokhale2022benchmarking,feng2023training,saharia2022photorealistic,cho2023dall,bakr2023hrs}, and some recent works~\citep{huang2024t2i,meng2024phybench} have also recognized the issue of factuality hallucination and introduced factuality benchmarks. Fig.~\ref{faithfulness} and Fig.~\ref{factuality} present examples of different kinds of T2I hallucinations.


\textbf{Faithfulness Hallucinations.}
We conduct the same taxonomy of faithfulness hallucinations in T2I tasks like I2T tasks: object-level, attribute-level and scene-level. Recent works primarily focus on object-level~\citep{cho2023dall} and attribute-level~\citep{Hu2023TIFAAA,feng2023training} hallucination, while some studies have explored scene-level hallucination, such as relationship hallucination~\citep{huang2023t2i,huang2024t2i}.

\begin{itemize}
    \item \textit{Object-level} hallucinations refer to cases where T2I models generate incorrect basic object information during image synthesis, including object type~\citep{Hu2023TIFAAA}, object number~\citep{cho2023dall} and so on. In Fig.~\ref{faithfulness}(a), the fruits in the generated image are pears rather than peaches in the text prompt.
    \item \textit{Attribute-level} hallucinations refer to cases where T2I models properly generate the object-level content, but the attributes of objects are wrong, such as color~\citep{feng2023training}, shape~\citep{huang2023t2i}, and so on, which have been explicitly specified in the text prompt. In Fig.~\ref{faithfulness}(b), the generated image depicts a triangular clock, which does not match the ``oval'' shape specified in the text prompt.
    \item \textit{Scene-level} hallucinations refer to T2I models incorrectly generating large-scale image content or inter-object information, including scene~\citep{bakr2023hrs}, object relationship~\citep{huang2023t2i,huang2024t2i} and so on. In Fig.~\ref{faithfulness}(c), the spatial relationship between the computer mouse and the book in the generated image is ``above'' rather than ``to the right of''. 
\end{itemize}

\textbf{Factuality Hallucinations.}
T2I models can generate realistic images according to text prompts, but the image content often contradicts established world knowledge. Recent works primarily focus on commonsense knowledge~\citep{huang2024t2i,niu2025wise}, such as the appearance of well-known landmarks or specific types of animals. Hallucinations involving domain-specific knowledge have received relatively little attention. Similar to VLMs, factuality hallucinations in T2I models can also be categorized into commonsense-based and domain-specific hallucinations.
\begin{itemize}
    \item \textit{Commonsense-based} hallucinations refer to cases where T2I models generate image content that conflicts with basic commonsense knowledge, such as the features of a dog or soccer ball~\citep{huang2024t2i}. Fig.~\ref{factuality}(a), the generated image depicts a mountain landscape, whereas the prompt is expected to evoke an image of the pyramids.
    \item \textit{Domain-specific} hallucinations arise when the image content produced by T2I models deviates from established domain knowledge, such as physical laws~\citep{niu2025wise}, scientific facts~\citep{lim2025evaluating} and so on. In Fig.~\ref{factuality}(b), the image generated by the T2I model not only fails to depict the refraction of the laser beam within the glass, but also shows an incorrect reflection angle on the glass surface. In Fig.~\ref{factuality}(c), the image presents an abdominal X-ray instead of the expected chest X-ray, indicating an inaccurate medical understanding.
\end{itemize}

\subsection{Relationship between Hallucination Problems in I2T and T2I tasks}
\label{subsec:hali2tt2i}
Building on the proposed definitional and classificatory framework, we can more clearly characterize the relationships and differences between hallucinations in I2T and T2I models. Under a unified definition, both faithfulness and factuality hallucinations emerge in both task settings, driven largely by shared factors such as cross-modal misalignment and insufficient or biased training data. Because faithfulness hallucinations stem from image-text alignment issues, I2T and T2I models exhibit similar patterns across object-, attribute-, and scene-level granularities.

Regarding factuality hallucinations, research has examined commonsense errors in both I2T and T2I tasks, while domain-specific hallucinations show clearer differences across model types. In I2T, studies increasingly focus on specialized domains such as medicine~\citep{chen2024detecting,seth2024hallucinogen,hartsock2024vision} and smart agriculture~\citep{arshad2025leveraging}, supported by domain-specific benchmarks and metrics. Many I2T models pretrained on large-scale medical datasets~\citep{ye2025multimodal} are evaluated using coarse-grained generative metrics (\emph{e.g.}, BERTScore~\citep{zhangbertscore}) or fine-grained discriminative metrics (\emph{e.g.}, accuracy, precision)~\citep{seth2024hallucinogen,hartsock2024vision}.  
In contrast, research on domain-specific hallucinations in T2I remains limited. Medical T2I models, typically GAN- or diffusion-based~\citep{uzunova2019multi,pinaya2022brain,polamreddy2024leapfrog}, are evaluated with general image quality metrics such as FID~\citep{heusel2017gans} and CLIPScore~\citep{hessel2021clipscore}. These metrics primarily measure overall image fidelity or semantic similarity but provide limited insight into fine-grained factual correctness~\citep{jayasumana2024rethinking, koishigarina2025clip}, especially in safety-critical domains. Overall, domain-specific evaluation is expanding for I2T, and developing fine-grained metrics for T2I represents a key direction for future research.

\begin{table*}[!t] 
\caption{Summary of representative evaluation metrics for hallucination in MLLMs.}
\setlength{\tabcolsep}{5mm}
\label{tab:hallucination-metrics}
\centering
\resizebox{0.93\linewidth}{!}{
\begin{tabular}{
    >{\centering\arraybackslash}m{0.15\linewidth}
    >{\centering\arraybackslash}m{0.17\linewidth}
    >{\centering\arraybackslash}m{0.30\linewidth}
    >{\centering\arraybackslash}m{0.36\linewidth}
}
\hline
\textbf{Metric} & \textbf{Formula} & \textbf{Usage} & \textbf{Derivatives} \\
\hline

Accuracy 
& $\dfrac{\text{TP} + \text{TN}}{\text{TP} + \text{FP} + \text{TN} + \text{FN}}$
& Accuracy measures the overall correctness of answers to YNQs or MCQs
& 
\makecell[c]{
Accuracy+ \\
\citep{fu2023mme}
}\\
\hline

Precision 
& $\dfrac{\text{TP}}{\text{TP} + \text{FP}}$
& Precision measures the proportion of predicted positives that are actually correct.
& --- \\
\hline

Recall 
& $\dfrac{\text{TP}}{\text{TP} + \text{FN}}$
& Recall measures the proportion of actual positives that are correctly identified.
& --- \\
\hline

F1 Score
& $ \dfrac{2 \times\text{Precision} \cdot \text{Recall}}{\text{Precision} + \text{Recall}}$
&
F1 Score measures the harmonic mean of precision and recall to balance false positives and false negatives.
&
\shortstack[c]{
$\text{F1}_{\text{PhD}} = 2 \cdot \dfrac{\text{Yes Recall} \times \text{No Recall}}{\text{Yes Recall} + \text{No Recall}}$ \\
\citep{liu2025phd} \\
$\text{Composite} = \text{MAE} /  \text{F1 Score}$ \\ \citep{heiman2025factchexcker}
} \rule{0pt}{60pt}
\\
\hline

CHAIR$_{\text{s}}$~\citep{rohrbach2018object} 
& $\dfrac{|\{\text{hall. sents.}\}|}{|\{\text{all sentences}\}|}$ 
& CHAIR$_{\text{s}}$ measures the proportion of sentences containing any hallucinated objects.
& 
\shortstack[c]{
Attack Success Rate (ASR)\\
\citep{wu2024autohallusion}\\ 
OpenCHAIR\\
~\citep{ben2024mitigating}} \rule{0pt}{47pt} \\
\hline

CHAIR$_{\text{i}}$~\citep{rohrbach2018object} 
& $\dfrac{|\{\text{hall. obj.}\}|}{|\{\text{all obj.}\}|}$ 
& CHAIR$_{\text{i}}$ measures the proportion of hallucinated objects among all mentioned objects.
& $\text{AMBER} = \dfrac{1}{2} \times (1 - \text{CHAIR} + \text{F1})$ \citep{wang2023amber}\\
\hline

GPT Rate~\citep{sun2024aligning} 
& 0--6 rating (6 = no hallucination) 
& GPT Rate uses GPT as an evaluator to assess the degree of hallucination.
& GAVIE~\citep{Liu2023MitigatingHI}, Fine-tuned model rate~\citep{wang2023evaluation}  
$\text{FaithScore} = \frac{\sum_{\text{i=1}}^{\text{C}}\sum_{\text{j=1}}^{\text{$\text{n}_\text{i}$}}\text{w}_\text{j}^\text{i} \cot \text{s}(\text{e}_\text{j}^\text{i},\text{I})}{\sum_{\text{i}=1}^{\text{C}}\text{n}_\text{i}}$ \citep{Jing2023FaithScoreFE} \\ 
\hline

Visor~\citep{gokhale2022benchmarking}
& $\dfrac{\text{Correct images}}{\text{Total images}}$
& Visor measures the proportion of images with correct spatial relationships.
& UniDet~\citep{huang2023t2i} \\
\hline

VQAScore~\citep{li2024evaluating}
& Alignment score
& VQAScore evaluates the alignment between image and text using a VQA model.
& 
\shortstack{
TIFA Score~\citep{Hu2023TIFAAA}\\
I-HallA Score~\citep{lim2025evaluating} \\ 
WiScore~\citep{niu2025wise}\\
LLMScore~\citep{lu2023llmscore} 
} \rule{0pt}{49pt}\\ 
\hline

\end{tabular}
}
\end{table*}

\begin{table*}[!t]
\caption{Summary of representative evaluation benchmarks for hallucination in MLLMs, with task types classified as discriminative (Dis), generative (Gen), and comprehensive (Comp) tasks.}
\label{benchmarks}
\centering
\resizebox{0.93\linewidth}{!}{
\setlength{\tabcolsep}{1mm} 
\renewcommand{\arraystretch}{1.1} 
\begin{tabular}{llcccccc} 
\hline
\multirow{2.5}{*}{\textbf{Task}} & \multirow{2.5}{*}{\textbf{Benchmark}} & \multirow{2.5}{*}{\textbf{Benchmark Type}} & \multirow{2.5}{*}{\textbf{Data Source}} & \multirow{2.5}{*}{\textbf{Size}} & \multicolumn{2}{c}{\textbf{Hallucination Type}} & \multirow{2.5}{*}{\textbf{Metric}} \\
\cmidrule(lr){6-7}
 & & & & & \textbf{Faithfulness} & \textbf{Factuality} & \\
\hline
\multirow{26}{*}{I2T} 
  & POPE~\citep{Li2023EvaluatingOH} & Dis & MSCOCO~\citep{lin2014microsoft} & 3000 & \cmark & \xmark & Acc/P/R/F1 \\
  & MME~\citep{fu2023mme} & Dis & MSCOCO~\citep{lin2014microsoft} & 1277 & \cmark & \cmark & Acc \\
  & CIEM~\citep{hu2023ciem} & Dis & MSCOCO~\citep{lin2014microsoft} & 72941 & \cmark & \xmark & Acc/P/R/F1 \\
  & RAH-Bench~\citep{chen2023mitigating} & Dis & MSCOCO~\citep{lin2014microsoft} & 3000 & \cmark & \xmark & FP \\
  & HallusionBench~\citep{Guan2023HallusionBenchAA} & Dis & Website & 1129 & \cmark & \cmark & Acc \\
  & FGHE~\citep{Wang2023MitigatingFH} & Dis & MSCOCO~\citep{lin2014microsoft} & 200 & \cmark & \xmark & Acc/P/R/F1 \\
  & ROPE~\citep{chen2024multi} & Dis & \makecell[c]{MSCOCO~\citep{lin2014microsoft} \\ \& ADE20K~\citep{zhou2017scene}} & 5000 & \cmark & \xmark & Acc \\
  & LongHalQA~\citep{qiu2024longhalqa} & Dis & \makecell[c]{Visual Genome~\citep{Krishna2016VisualGC} \\ \& Object365~\citep{shao2019objects365}}
   & 6485 & \cmark & \xmark & Acc \\
  & R-Bench~\citep{Wu2024EvaluatingAA} & Dis & MSCOCO~\citep{lin2014microsoft} & 11651 & \cmark & \xmark & Acc/P/R/F1 \\
  & VisDiaHalBench~\citep{cao2024visdiahalbench} & Dis & GQA~\citep{hudson2019gqa} & 25000 & \cmark & \xmark & F1/EM \\
  & AutoHallusion~\citep{wu2024autohallusion} & Dis & MSCOCO~\citep{lin2014microsoft} \& DALL-E-2 & 1000 & \cmark & \xmark & ASR/MASR/CASR \\
  & PhD~\citep{liu2025phd} & Dis & TDIUC~\citep{kafle2017analysis} & 102564 & \cmark & \cmark & $\text{F1}_{\text{PhD}}$ \\
  \cmidrule(lr){2-8}
  & CHAIR~\citep{rohrbach2018object} & Gen & MSCOCO~\citep{lin2014microsoft} & 5000 & \cmark & \xmark & CHAIR \\
  & CC-Eval~\citep{zhai2023halle} & Gen & Visual Genome~\citep{Krishna2016VisualGC} & 100 & \cmark & \xmark & CHAIR/Cover \\
  & HaELM~\citep{wang2023evaluation} & Gen & MSCOCO~\citep{lin2014microsoft} & 5000 & \cmark & \xmark & LLM Rate \\
  & GAVIE~\citep{Liu2023MitigatingHI} & Gen & Visual Genome~\citep{Krishna2016VisualGC} & 1000 & \cmark & \xmark &  GPT Rate \\
  & OpenCHAIR~\citep{ben2024mitigating} & Gen & Stable Diffusion & 5000 & \cmark & \xmark & OpenCHAIR \\
  & MMHal-Bench~\citep{sun2024aligning} & Gen & OpenImages~\citep{kuznetsova2020open} & 96 & \cmark & \xmark & GPT Rate \\
  & NOPE~\citep{lovenia2024negative} & Gen & OpenImages~\citep{kuznetsova2020open} & 36000 & \cmark & \xmark & Acc/METEOR \\
  & VHILT~\citep{rani2024visual} & Gen & Website & 2000 & \cmark & \cmark & Acc \\
  & HallucinaGen~\citep{seth2024hallucinogen} & Gen & \makecell[c]{MSCOCO~\citep{lin2014microsoft} \\ \& NIH Chest X-ray~\citep{wang2017chestx}} & 96000 & \cmark & \cmark & Acc \\
  \cmidrule(lr){2-8}
  & AMBER~\citep{wang2023amber} & Comp & Website & 15202 & \cmark & \xmark & Acc/CHAIR \\
  & VHtest~\citep{huang2024visual} & Comp & MSCOCO~\citep{lin2014microsoft} \& DALL-E-3 & 1200 & \cmark & \xmark & Acc \\
  & Hal-Eval~\citep{Jiang2024HalEvalAU} & Comp & MSCOCO~\citep{lin2014microsoft} & 10000 & \cmark & \xmark & Acc/Score \\
  & Med-HallMark~\citep{chen2024detecting} & Comp & Slake~\citep{liu2021slake}, etc. &  7341 & \cmark & \cmark & MediHall Score\\
  & MERLIM~\citep{villa2025behind} & Comp & MSCOCO~\citep{lin2014microsoft} & 31373 & \cmark & \xmark & Acc/F1 \\
  & ODE~\citep{tu2025ode} & Comp & Stable Diffusion & 8786 & \cmark & \cmark & AMBER/Acc \\
\hline
\multirow{12}{*}{T2I}
  &$\text{SR}_{\text{2D}}$~\citep{gokhale2022benchmarking} & Gen & MSCOCO~\citep{lin2014microsoft} & 25280 & \cmark & \xmark & Visor Score \\
  &DrawBench~\citep{saharia2022photorealistic} & Gen & Human, DALL-E & 200 & \cmark & \xmark & Human\\
  & ABC-6K \& CC-500~\citep{feng2023training} & Gen & MSCOCO~\citep{lin2014microsoft} & 6900 & \cmark & \xmark & Human/CLIP-R \\ 
  & PaintSkills~\citep{cho2023dall} & Gen & Template & 7330 & \cmark & \xmark & DETR Score\\  
  & HRS-Bench~\citep{bakr2023hrs} & Gen & GPT & 15000 & \cmark & \xmark & P/AC-T2I \\
  & T2I-CompBench~\citep{huang2023t2i} & Gen & \makecell[c]{ MSCOCO~\citep{lin2014microsoft} \\ \& Template \& GPT} & 6000 & \cmark & \xmark & BLIP-VQA/UniDet \\
  & TIFA v1.0~\citep{Hu2023TIFAAA} & Gen & MSCOCO~\citep{lin2014microsoft} & 4000 & \cmark & \xmark & TIFA Score \\
  & T2I-FactualBench~\citep{huang2024t2i} & Gen & GPT & 3000 & \cmark & \cmark & VQA Score \\
  & GenAI-Bench~\citep{li2024evaluating} & Gen & Human & 1600 & \cmark & \xmark & Human/VQAScore \\
  & I-HallA v1.0~\citep{lim2025evaluating} & Gen & Textbook & 200 & \cmark & \cmark & I-HallA Score \\
  & WISE~\citep{niu2025wise} & Gen & LLM-Constructed & 1000 & \cmark & \cmark & WiScore \\
\hline
\end{tabular}
}

\end{table*}

\section{Hallucination Evaluation Benchmarks and Metrics}
\label{sec:benchmark}
In this chapter, we provide a systematic overview of the evolution of existing benchmarks. Tab.~\ref{tab:hallucination-metrics} summarizes metrics for hallucination evaluation, including their formula and derivatives. Tab.~\ref{benchmarks} summarizes key details of these benchmarks, including their data sources, construction paradigms, and evaluation criteria. Hallucination benchmarks for I2T and T2I models are introduced in Sec.~\ref{beni2t} and Sec.~\ref{bent2i} respectively, with their interrelation explored in Sec.~\ref{beni2tt2i}.

\subsection{Hallucination Benchmarks for I2T Models}
\label{beni2t}
Effectively evaluating hallucination in VLMs remains an open challenge. Recent I2T benchmarks assess hallucination across diverse task formats, including Yes or No Questions (YNQs), Multiple-Choice Questions (MCQs), Image Captioning (IC), and Visual Question Answering (VQA). These tasks are typically categorized by response type into two groups: discriminative (YNQs and MCQs) and generative (IC and VQA)~\citep{bai2024hallucinationsurvey,lan2024survey}. Following this convention, we categorize hallucination benchmarks based on whether they constrain the output space (discriminative) or allow open-ended generation (generative).\par

In addition, some benchmarks integrate both discriminative and generative tasks to analyze the inconsistencies in hallucination behavior across different evaluation tasks~\citep{wang2023amber,Jiang2024HalEvalAU}, which has not been sufficiently addressed in previous surveys~\citep{bai2024hallucinationsurvey,lan2024survey}. To capture this feature, we introduce a third category, namely comprehensive benchmarks. This classification provides a clearer framework for understanding the design and focus of existing hallucination evaluations. We present representative benchmarks within each category in the following sections.\par

\subsubsection{\textbf{Discriminative Benchmarks}}

\label{i2tdis}
A growing body of benchmarks adopts discriminative tasks for hallucination evaluation. Compared with generative settings, VLMs tend to show more stable performance on discriminative tasks~\citep{Li2023EvaluatingOH}, which also enable fine-grained diagnosis of different hallucination types via diverse YNQs and MCQs. As more hallucination categories emerge and LLMs gain stronger in-context learning abilities, discriminative benchmarks have expanded beyond object hallucinations, with increasingly automated construction pipelines. Considering these developments, we summarize discriminative benchmarks along two axes: their coverage of hallucination types and their data generation strategies, highlighting differences in data sources, construction methods, and evaluation metrics.\par

POPE~\citep{Li2023EvaluatingOH} is an early YNQ-based benchmark for assessing object hallucinations in VLMs. It selects 500 MSCOCO~\citep{lin2014microsoft} validation images containing at least three annotated 
objects and constructs questions of the form ``Is there a \{object\} in the image?''. For ``Yes'' queries, annotated objects are used; for ``No'' queries, non-existent objects are sampled via three strategies: 
random sampling, popular sampling (common objects), and adversarial sampling (frequent co-occurrence objects). Evaluation metrics include Accuracy, Precision, and Recall. Model performance degrades from 
random to popular and adversarial settings, reflecting training data biases. While POPE is widely used for evaluating basic object-existence hallucinations, its closed-form YNQs and single-object focus 
limit its alignment with open-ended real-world scenarios.\par

ROPE~\citep{chen2024multi} extends hallucination evaluation in VLMs from single- to multi-object scenarios using MCQs. Models must identify correct object names from a list across multiple bounding boxes, making the task more challenging than POPE. Images are drawn from MSCOCO~\citep{lin2014microsoft} and ADE20K~\citep{zhou2017scene}, focusing on those with more than five annotated objects and bounding boxes with IoU below 0.1. The benchmark is divided into four subsets: homogeneous, heterogeneous, adversarial, and in-the-wild. Results show that hallucination likelihood increases with the number of object categories, indicating that object co-occurrence bias remains a key driver of I2T hallucinations. ROPE has become a standard benchmark for multi-object hallucination analysis, with its dense-object and co-occurrence settings better reflecting complex real-world scenes.\par

FGHE~\citep{Wang2023MitigatingFH} builds upon the POPE framework and extends the scope of hallucination evaluation to include not only object hallucination, but also relationship, attribute, 
and behavior hallucinations using YNQs. This benchmark selects 50 images from MSCOCO~\citep{lin2014microsoft}, and human annotators construct 200 binary YNQs based on these images. Evaluation is 
performed using standard metrics such as Accuracy, Precision, F1 score, and other relevant indicators, enabling a more comprehensive analysis of fine-grained hallucination types in VLMs.\par

MME~\citep{fu2023mme} evaluates VLMs from perceptual and cognitive perspectives across various hallucination types. Perception tasks include coarse-grained recognition of objects, quantity, 
color, and location, and fine-grained tasks requiring external knowledge, such as movie posters, celebrity photos, and OCR. Cognition tasks cover commonsense reasoning, math, translation, 
and code. Images are drawn from public datasets like MSCOCO~\citep{lin2014microsoft}, and YNQs are manually constructed. The Accuracy+ metric based on Yes-No paired questions mitigates 
Yes/No bias, offering a reliable assessment. MME's broad coverage approximates real-world multimodal demands and remains widely used.\par

HallusionBench~\citep{Guan2023HallusionBenchAA} is a comprehensive benchmark evaluating two hallucination types based on visual necessity: Visual Dependent (requiring image analysis for reasoning) and Visual Supplement (reliant on external knowledge). By utilizing manually edited images and a Yes/No bias test, the benchmark assesses model robustness and quantitative preferences. Results indicate that even advanced VLMs struggle (\emph{e.g.}, GPT-4V achieving only 31.42\% accuracy) revealing a significant tendency for models to prioritize internal parametric memory over actual visual context.''\par

CIEM~\citep{hu2023ciem} targets hallucinations regarding object existence, attributes, and relationships through an automated benchmark construction pipeline. By leveraging MSCOCO~\citep{lin2014microsoft} annotations and GPT-3.5, it generated over 72,000 balanced YNQs, achieving a scale significantly surpassing manual benchmarks. While this method relies on pre-existing annotations, limiting its application to raw images, human verification confirmed a low 5\% error rate, validating the efficacy of LLM-driven generation.\par

RAH-Bench~\citep{chen2023mitigating} evaluates three types of hallucinations: object classification, attributes, and relationships. This work collects images from the MSCOCO validation set 
and utilizes GPT-4 to generate 3000 YNQs based on annotations. Unlike works~\citep{Li2023EvaluatingOH,hu2023ciem} evaluating all YNQs using the Accuracy metric, RAH-Bench introduces False 
Positive (FP) Rates to assess the model's performance across different hallucination types. While Accuracy reflects overall correctness, False Positive Rates provide a more fine-grained 
diagnostic signal. Experiments show that VLMs perform worse in attribute and relationship hallucinations, suggesting targeted improvements.\par

R-Bench~\citep{Wu2024EvaluatingAA} addresses the scarcity of relationship hallucination evaluations by introducing a large-scale benchmark (11,651 instances) covering existence, spatial relationships, and actions. It utilizes a hierarchical structure with image-level (existence) and instance-level (specific types) to
to enable fine-grained analysis. Constructed using scene graph parsers on MSCOCO~\citep{lin2014microsoft} and LLM-generated YNQs based on Nocaps~\citep{agrawal2019nocaps}, R-Bench highlights the specific vulnerability of VLMs to relational reasoning errors.\par

AutoHallusion~\citep{wu2024autohallusion} is an automated benchmark pipeline for evaluating object existence and spatial relationship hallucinations in VLMs. It uses text prompts to probe 
the model's language priors (\emph{e.g.}, ``We have \{object1\} in the image. What is the object to insert/remove under the strategy X where we wish to...'') and performs targeted image 
editing based on the responses. The pipeline employs three induction strategies: abnormal object insertion, paired object insertion, and correlated object removal. Hallucination detection 
with GPT-4V-Turbo achieves 92.6\% accuracy. Experiments show induction success rates of 97.7\% on synthetic datasets and 98.7\% on real-world datasets. AutoHallusion demonstrates that 
exploiting VLMs' language priors can amplify hallucination rates and provides a scalable method to systematically probe datasets for induced hallucinations.\par

PhD~\citep{liu2025phd} evaluates hallucinations in VLMs across object existence, attributes, and counting. It uses images and annotations from TDIUC~\citep{kafle2017analysis} and 
generates word sets for different question types. For color questions, GPT creates alternative color words excluding the correct one, and CLIP measures similarity between the image 
and descriptions. GPT then generates binary YNQ questions based on the highest similarity keyword, increasing difficulty compared to direct GPT question generation. PhD also employs 
hallucination induction via inaccurate prefixes and counterfactual images. To mitigate bias from models' ``Yes'' or ``No'' preference, a new $\text{F1}_{\text{PhD}}$ score is 
adopted (as it shown in Tab.~\ref{tab:hallucination-metrics}). PhD emphasizes challenging question design, induction methods and careful evaluation metrics for hallucination evaluation.\par

LongHalQA~\citep{qiu2024longhalqa} is a benchmark of 6,000 long texts covering hallucinations in object existence, attributes, counting, and relationships. Unlike short-context 
questions in previous benchmarks, it introduces object-level description, image-level description, and multi-round conversation formats, providing rich contexts and reflecting 
real-world interactions. GPT-4V generates MCQs for both discriminative and generative tasks: description selection and continuation selection. Results show model rankings in 
LongHalQA align with free-generation settings, demonstrating its effectiveness. LongHalQA highlights VLMs' limitations in reasoning and grounding over extended text and offers 
a unified framework for LLM-free evaluation.\par

\textbf{Summary of Discriminative Benchmarks.}
As shown in Tab.~\ref{benchmarks}, discriminative benchmarks show clear trends in evaluation focus and construction methods. The focus has shifted from basic faithfulness hallucinations to factuality hallucinations involving commonsense and domain knowledge, moving from perceptual to cognitive assessment. Early benchmarks rely on manually designed questions, while recent efforts such as CIEM~\citep{hu2023ciem} and RAH-Bench~\citep{chen2023mitigating} use LLMs for automated generation. Frameworks like AutoHallusion~\citep{wu2024autohallusion} and PhD~\citep{liu2025phd} enable automated pipelines and diverse hallucination induction methods. LongHalQA~\citep{qiu2024longhalqa} unifies generative and discriminative tasks via MCQs, a creative benchmark construction method.\par

Currently, research on factuality hallucinations in discriminative benchmarks remains limited, with a predominant focus mainly on commonsense knowledge and insufficient coverage of specialized domains. To better align with real-world application scenarios, future discriminative benchmark construction can incorporate richer contextual formats such as multi-turn dialogues and long-text. Additionally, exploring more hallucination induction methods similar to PhD~\citep{liu2025phd} could also enhance the complexity of discriminative benchmarks.

\subsubsection{\textbf{Generative Benchmarks}}

\label{i2tgen}
Generative tasks produce longer, open-ended responses, challenging traditional accuracy-based metrics and requiring dedicated hallucination detection. Early work focused on object hallucinations in image captioning~\citep{rohrbach2018object}, while recent benchmarks leverage large language models for contextual understanding and long-form generation~\citep{ben2024mitigating,zhai2023halle}. Concerns over data leakage~\citep{rani2024visual,sun2024aligning} have prompted the use of alternative data sources, such as website-crawled or synthetic images~\citep{rani2024visual,ben2024mitigating}, to improve evaluation robustness. This section reviews generative hallucination benchmarks, highlighting diverse hallucination types, expanded task formats, and automated detection methods, with attention to question types, data sources, and evaluation strategies.\par

CHAIR (Caption Hallucination Assessment with Image Relevance)~\citep{rohrbach2018object} is one of the early works evaluating object hallucination in the image captioning task. 
It measures the proportion of hallucinated objects in VLM responses with two variants: per-instance ($\text{CHAIR}_\text{i}$) and per-sentence ($\text{CHAIR}_\text{s}$), as shown in Tab.~\ref{tab:hallucination-metrics}. Detection uses sentence tokenization and synonym mapping, and analysis is limited to the 80 MSCOCO~\citep{lin2014microsoft} object types.\par

OpenCHAIR~\citep{ben2024mitigating}is an open-vocabulary benchmark designed to evaluate object hallucination in the image captioning task. Compared to CHAIR~\citep{rohrbach2018object}, OpenCHAIR expands the evaluation scope to include a wider range of rare and diverse objects, reflecting the variability of real-world scenarios. To construct the benchmark, LLM is used to generate diverse captions, and then a diffusion model generates the corresponding images. Additionally, OpenCHAIR replaces CHAIR's fixed synonym lists with a dynamic semantic matching method based on LLMs, enabling open-vocabulary and flexible object hallucination evaluation.

CC-Eval~\citep{zhai2023halle} evaluates object hallucination in detailed image captions using GPT-4 assistance. It uses 100 randomly sampled images from Visual Genome~\citep{Krishna2016VisualGC}, and requires VLMs to generate captions. GPT-4 extracts mentioned objects and compares them with the ground-truth annotations from Visual Genome. The benchmark adopts CHAIR to measure hallucination rate and introduces coverage to assess how well real objects are captured. These metrics jointly offer a comprehensive assessment of object-level accuracy.

NOPE~\citep{lovenia2024negative} evaluates object hallucinations in cases where the answer should be ``None'' or ``Nothing'' (NegP data). To address the scarcity of such cases, it constructs 36,000 examples using two LLM-based methods: generate-from-scratch and list-then-rewrite. Generate-from-scratch prompts the LLM directly, while list-then-rewrite uses multi-turn dialogue. Human evaluation shows 50\% accuracy for the former and 92\% for the latter. NOPE employs Accuracy and METEOR, revealing that VLMs are prone to hallucinations on non-existent objects, reflecting a loss of negative data during training.\par

VHILT~\citep{rani2024visual} collects 2,000 images from the New York Times Twitter account, with VQA and caption tasks manually designed via Amazon Mechanical Turk. It defines eight hallucination types, including contextual guessing, identity incongruity, and geographic erratum. Evaluation relies on human annotations, showing models frequently hallucinate on contextual guessing and identity incongruity, highlighting limitations in visual cognition and reasoning. Due to its manual annotation dependence, future work could explore using large models to automate parts of the benchmark.\par

MMHal-Bench~\citep{sun2024aligning} collects images from the validation set of OpenImages~\citep{kuznetsova2020open} to prevent data leakage. The benchmark contains 96 handcrafted VQAs, and covers 8 types of hallucinations, including object existence, attributes, counting, and environment. Evaluation leverages GPT-4 to analyze and score the degree of hallucination based on the consistency with actual image content. Results show that GPT-4’s hallucination scores align with human annotations in 94\% of cases, demonstrating the effectiveness of using LLMs for evaluation.\par

HaELM~\citep{wang2023evaluation} evaluates VLMs' performance on image captioning tasks utilizing a fine-tuned LLM. To construct the training dataset for fine-tuning, GPT is prompted to create 10,000 hallucinatory and 10,000 non-hallucinatory captions based on the MSCOCO~\citep{lin2014microsoft} training set. LLaMA is then fine-tuned with LoRA to create a hallucination detector. For evaluation, 5,000 samples from MSCOCO test set are used. Results show that HaELM achieves 95\% of GPT's accuracy in hallucination detection. HaELM provides a scalable and cost-efficient solution without dependence on external APIs.\par

HallucinaGen~\citep{seth2024hallucinogen} evaluates VLMs' object recognition and reasoning capabilities through challenging visual-language tasks with increasing difficulty: Localization (LOC), Visual Context (VC), and Counterfactual (CF). Images are sourced from MSCOCO~\citep{lin2014microsoft} and NIH Chest X-ray~\citep{wang2017chestx}. Similar to NOPE~\citep{lovenia2024negative}, it employs implicit hallucination induction attacks, including locating absent entities and imagining non-existent objects. Evaluation combines word-matching algorithms with GPT-4o, using five prompts for robust assessment. Results show that as task difficulty increases, VLMs exhibit higher hallucination rates. In tasks involving domain-specific knowledge, even specialized models like LLaVA-Med struggles to perform better than random guessing. This highlights the vulnerability of VLMs in high-stakes scenarios like healthcare, where accuracy and reliability are critical.\par

\textbf{Summary of Generative Benchmarks.} Current generative benchmarks have advanced in diversifying hallucination types, expanding task formats, and introducing automation. However, several challenges remain. Many benchmarks still rely on coarse metrics such as hallucination rates and LLM scores, lacking fine-grained assessment. Data leakage risks persist even with web-crawled or validation-set data. Automation is mostly limited to question generation and depends on human verification. Developing scalable pipelines that combine reliable automation with diagnostic evaluation is a key direction.

\subsubsection{\textbf{Comprehensive Benchmarks}}

\label{i2tcom}
Considering the limited scope of discriminative benchmarks and the challenges of generative benchmarks, comprehensive benchmarks combine both approaches for a more holistic evaluation. We review existing benchmarks focusing on three aspects: broader hallucination types, more automated and scalable evaluation methods, and mitigation of data leakage risks. As shown in Tab.~\ref{tab:hallucination-metrics} and Tab.~\ref{benchmarks}, each benchmark is summarized in terms of hallucination types, evaluation metrics, data sources, and core design features.

MERLIM~\citep{villa2025behind} evaluates existence, relation, and counting hallucinations using a large number of both edited and original images. GPT is used for automatic question generation. To construct the benchmark, the original images are edited by removing a selected object and seamlessly inpainting the missing region to preserve visual realism, creating paired edited versions for controlled comparison. By comparing model responses between the original and edited images, hallucinations lacking visual grounding can be effectively identified. Relationship and counting hallucinations are evaluated via YNQs. Object hallucinations are evaluated via open-ended questions, while model responses are processed with spaCy library and GPT, followed by calculating metrics including Precision.\par
AMBER~\citep{wang2023amber} evaluates hallucinations in object existence, attributes, and relations through discriminative and generative tasks. Images are manually collected to prevent data leakage. Discriminative questions reveal that models like GPT-4V and Qwen-VL favor ``No'' answers, while generative tasks extract target objects from captions and compute metrics such as CHAIR, though limited by extraction errors and focus on existence hallucinations. Unlike MERLIM, AMBER introduces a unified AMBER Score (as shown in Tab.~\ref{tab:hallucination-metrics}) to assess overall performance, providing a holistic benchmark for model reliability across tasks.\par

VHtest~\citep{huang2024visual} is a synthetic benchmark designed to evaluate eight types of visual hallucinations, including object existence, shape, color, and OCR, across 1,200 instances. To ensure image diversity and prevent data leakage, it constructs the dataset by first identifying initial hallucination instances from MSCOCO~\citep{lin2014microsoft} using CLIP and DINO, and then generating additional images with T2I models based on annotations. Question construction utilizes a combination of templates and manual editing to create YNQs and image captioning tasks. YNQ responses are automatically evaluated, while captions require manual assessment. Experimental results show that this benchmark effectively reduces the accuracy of state-of-the-art models like GPT-4V and LLaVA-1.5.

VisDiaHalBench~\citep{cao2024visdiahalbench} includes 25,000 multi-turn questions and 5,000 edited images, covering hallucinations such as object existence, 
attributes, relations, and verification questions. Built on GQA~\citep{hudson2019gqa}, it edits images and scene graphs by removing, replacing, or changing 
object colors. The modified scene graphs are used to prompt GPT-4 to generate five-turn dialogues, enabling in-depth evaluation of VLMs' visual understanding 
and consistency. Evaluation applies keyword-filtered outputs to compute macro-average F1 and Exact Match scores. Results show GPT-4 performs significantly 
worse on VisDiaHalBench, allowing fine-grained analysis of reasoning and consistency errors that single-turn evaluations could miss.\par

Hal-Eval~\citep{Jiang2024HalEvalAU} provides a systematic quantitative analysis of data leakage in VLMs. It constructs evaluation datasets from MSCOCO~\citep{lin2014microsoft} (in-domain) and various web sources (out-of-domain) and generates questions with GPT-4. Hal-Eval covers discriminative and generative tasks, addressing hallucinations in existence, attributes, relations, and event types. Experiments show most models favor ``Yes'' in discriminative tasks, revealing a bias toward positive responses. Models like LLaVA exhibit stronger hallucinations out-of-domain, likely due to extensive in-domain fine-tuning. Generative task evaluation uses a fine-tuned Hal-Evaluator, performing comparably to GPT-4 and effectively identifying hallucinations across datasets.\par

Med-HallMark~\citep{chen2024detecting} is a hallucination detection benchmark for the multi-modal medical domain, supporting Med-VQA and Imaging Report Generation 
tasks. It evaluates model robustness using four question types: conventional, confidence-weakening, counterfactual, and image depiction, constructed via manual 
annotation, model-generated responses from LLaVA-Med and GPT, and prompt engineering. To characterize medical hallucinations, it introduces a hierarchical categorization 
with five levels—catastrophic, critical, attribute, prompt-induced, and minor—based on clinical impact. The MediHall Score quantifies hallucination severity at the answer level 
for Med-VQA and the sentence level for IRG. Med-HallMark provides structured metrics for rigorous safety assessment and supports systematic evaluation of multimodal models 
to enhance reliability in clinical applications.\par

ODE (Open-set Dynamic Evaluation)~\citep{tu2025ode} is a dynamic benchmark for assessing hallucinations in VLMs by generating novel image-text pairs with diverse objects and attributes. It represents objects and their co-occurrence using a weighted graph and produces test samples across standard, long-tail, random, and fictional distributions. This design mitigates data leakage and allows iterative updates to expand evaluation scope. ODE employs generative and discriminative tasks with specific inquiry templates, and evaluates performance using accuracy, CHAIR, and AMBER. Experiments show ODE effectively reveals hallucination patterns and supports fine-tuning to enhance model reliability.\par

\textbf{Summary of Comprehensive Benchmarks.}
As shown in the analysis and Tab.~\ref{benchmarks}, comprehensive benchmarks remain relatively scarce. While efforts like MERLIM~\citep{villa2025behind} and Hal-Eval~\citep{Jiang2024HalEvalAU} incorporate both task types, fully unified evaluation frameworks that consistently integrate results across different formats are still in progress. Benchmarks such as AMBER~\citep{wang2023amber} introduce unified metrics, though achieving balanced coverage of hallucination types across tasks remains a challenge. Nevertheless, several benchmarks provide valuable strategies for mitigating data leakage. For instance, Hal-Eval~\citep{Jiang2024HalEvalAU} includes both in-domain and out-of-domain datasets to assess potential data leakage. VHtest~\citep{huang2024visual} and ODE~\citep{tu2025ode} utilize T2I generation models to construct synthetic datasets, improving scalability and reducing the risk of data leakage. These developments underscore the need for more comprehensive benchmarks.

\textbf{Summary of I2T Hallucination Benchmarks.}
Our review of I2T hallucination benchmarks reveals several trends. The range of hallucination types is expanding, LLMs are increasingly used in benchmark construction, and data leakage concerns are receiving more attention. However, despite these advancements, notable imbalances remain. Benchmarks focusing on faithfulness are more developed, while those targeting factuality hallucinations remain limited (fewer than 5 I2T benchmarks in Tab.~\ref{benchmarks}). Research continues to emphasize image-text consistency, leaving broader factuality issues underexplored.\par

As shown in Tab.~\ref{benchmarks}, discriminative benchmarks outnumber generative ones, as predefined answer sets simplify hallucination detection. Generative benchmarks are less common due to the difficulty of assessing open-ended outputs. Similarly, most benchmarks focus on a single task, with fewer supporting both discriminative and generative evaluations. Developing comprehensive benchmarks is an important future direction. Moreover, current benchmarks mainly focus on single-turn tasks including YNQs or image captioning, with fewer benchmarks addressing complex tasks like multi-turn dialogue~\citep{cao2024visdiahalbench}. Future benchmarks could explore richer interaction formats to better reflect real-world applications.\par

\textbf{Analysis of I2T Hallucination Causes.}
Understanding why hallucinations occur in I2T models requires dissecting the contributing factors beyond merely measuring them. 
We attribute I2T hallucinations to two primary sources: (1) \textit{Data Limitations}, which include insufficient quantity, low diversity, noise, and bias in the training data; and (2) \textit{Model-Specific Mechanisms}, which encompass architectural constraints, attention dynamics, and cross-modal misalignment. 
This structure highlights how both the inputs and the model design jointly shape hallucination patterns, providing a systematic lens for analysis.

\begin{itemize}

\item \textit{Lack of Data.}
Insufficient quantity of training data remains a fundamental cause of hallucinations in I2T models. \cite{udandarao2024no} show that 
current datasets leave vast semantic spaces uncovered due to the exponential scaling required for zero-shot mastery.
This sparsity is compounded by distributional imbalances: models struggle with underrepresented long-tail concept~\citep{parashar2024neglected} and unseen
multi-object combinations~\citep{chen2024multi}, while a lack of negative samples impedes negation handling~\citep{lovenia2024negative}.
These findings suggest that data sparsity limits the model's ability to ground descriptions in visual evidence, pushing it to fall back on 
language priors, increasing the likelihood of hallucination.

\item \textit{Noisy Data.}
Beyond scarcity, the quality of I2T training data is compromised by inherent noise in both web-scraped and synthetic datasets.
\cite{yu2024hallucidoctor} identify a cascade of errors where LLM-generated instruction data contains systematic hallucinations,
frequently fabricating objects, relations and attributes. When models are fine-tuned on this corrupted supervision,
they inevitably internalize these hallucinatory patterns. These kinds of noisy data directly degrades the models' ability 
to ground text in visual evidence, leading to persistent factual inconsistencies in downstream generation.

\item \textit{Bias in Data.}
Biases in training data, such as frequency imbalance and spurious co-occurrence, also contribute significantly to hallucinations. 
Models systematically over-predict dominant classes (\emph{e.g.}, human), a tendency confirmed by the high false-positive rates~\citep{dai2023plausible,Li2023EvaluatingOH}.
Additionally, POPE~\citep{Li2023EvaluatingOH} finds that high-frequency objects increase hallucination rates in negative-sample tests.
Rigid co-occurrence patterns also override visual evidence~\citep{leng2024mitigating}, \emph{e.g.}, labeling a black banana as yellow.
These failures indicate a fundamental reliance on statistical shortcuts rather than faithful visual grounding, causing models to hallucinate plausible but absent details when inputs deviate from training norms of I2T models.

\end{itemize}

\begin{itemize}

\item \textit{Model Architecture.}
The architecture of I2T models, including visual encoders and attention mechanisms, plays a crucial role in hallucination 
generation. Perceptual bottlenecks in visual backbones (\emph{e.g.}, coarse resolution in CLIP-ViT/16) cause aliasing, where distinct details 
are compressed into ambiguous embeddings~\citep{Radford2021LearningTV,tong2024eyes,wei2024vary}. Attention misallocation prevents effective 
utilization of these features. Models frequently underutilize visual data even when available (POPEv2, \cite{li2025analyzing}) and fail to 
ground generated tokens in relevant image regions~\citep{xie2025tarac}. Autoregressive decoding process further amplifies 
these initial slips: as generation progresses, early grounding errors accumulate, causing the model to drift further from visual evidence in 
longer sequences~\citep{zhou2024analyzing}.

\item \textit{Cross-modal Misalignment.}
Hallucinations frequently stem from the fundamental difficulty in synchronizing disparate modalities~\citep{wu2025mitigating}. 
Misalignment stems from distributional gaps in the joint space~\citep{liu2025unraveling} that lead to erroneous object-attribute 
associations~\citep{alonso2025vision}. Furthermore, models exhibit a structural bias toward suppressing visual influence and 
causing the model to hallucinate based on language probabilities rather than image~\citep{zheng2025modality,huang2024opera,zhou2024analyzing,Guan2023HallusionBenchAA}. 
While recent works propose enhancing visual salience~\citep{xie2024v}, modulating textual signals~\citep{li2025treble}, or cross-modal verification~\citep{huang2024opera}, 
misalignment remain a key structural source of hallucination.

\end{itemize}

\subsection{Hallucination Benchmarks for T2I Models}

\label{bent2i}
While I2T benchmarks have advanced rapidly, hallucination evaluation in T2I models has also progressed in hallucination types, benchmark construction, and evaluation methods. Early works focused on image-text alignment, such as color~\citep{feng2023training} and spatial relationships~\citep{gokhale2022benchmarking}, categorized as faithfulness hallucinations. Recent studies~\citep{huang2024t2i} extend to factuality hallucinations using prompts based on world knowledge. Benchmark construction has become more automated, evolving from manual dataset sampling~\citep{feng2023training} to template-based or LLM-assisted generation~\citep{huang2023t2i,bakr2023hrs}. Evaluation methods have similarly shifted from manual annotation~\citep{saharia2022photorealistic} and object detection~\citep{gokhale2022benchmarking} to automated approaches using VLMs, VQA, and captioning tasks~\citep{huang2023t2i,Hu2023TIFAAA}, enabling fine-grained and scalable assessments.\par

The following section provides an overview of existing T2I hallucination benchmarks, highlighting hallucination types, data sources, and evaluation metrics. As shown in Tab.~\ref{benchmarks}, T2I benchmarks are generally generative, assessing image generation correctness.

$\text{SR}_{\text{2D}}$~\citep{gokhale2022benchmarking} is a benchmark designed to evaluate spatial hallucination in T2I models. Based on 80 object categories from MSCOCO~\citep{lin2014microsoft}, the benchmark contains a total of 25,280 text prompts. The VISOR metric is used to calculate the accuracy of the generated images. It extracts spatial relationships from generated images using object detection and centroid determination algorithms. Experimental results show that T2I models exhibit poor performance in understanding and rendering spatial relationships, as indicated by low VISOR scores.\par

ABC-6K \& CC-500~\citep{feng2023training} are two benchmarks targeting color hallucination in T2I models. ABC-6K constructs 6400 text prompts by sampling descriptions from MSCOCO~\citep{lin2014microsoft} and swapping color words in those that contain at least two color terms. CC-500 focuses on simple parallel text prompts, typically in the format of ``a red apple and a yellow banana.'' The evaluation includes human evaluation, automatic evaluation using the phrase alignment model GLIP, and system-level metrics such as FID~\citep{heusel2017gans}. Experimental results reveal three main challenges in generating images from combined prompts: attribute leakage, interchanged attributes, and missing objects.\par

DrawBench~\citep{saharia2022photorealistic} is a comprehensive benchmark including a diverse set of 200 prompts across 11 categories. These categories test various aspects of model performance, including color, quantity, spatial relationships, scene text, and unusual object interactions, as well as complex prompts with long descriptions, rare words, and misspellings. Human evaluations are conducted by presenting images from different models and asking raters to compare them in terms of image fidelity and image-text alignment.\par

PaintSkills~\citep{cho2023dall} assesses image-text alignment in T2I tasks using 7330 template-generated text prompts. It evaluates hallucinations such as object existence, counting, and spatial relationships. Text prompts are generated through templates, and over 50000 images are created using Unity engine to train an object detector (DETR). The evaluation employs DETR to automatically verify object presence in the generated outputs, and determines spatial relationships through algorithmic analysis. Results show that T2I models generally perform poorly in object counting and spatial relationships.\par

HRS-Bench~\citep{bakr2023hrs} is a scalable evaluation benchmark for T2I models, assessing accuracy, robustness, generalization, fairness, and bias. It covers 50 diverse scenes across domains such as fashion, animals, transportation, and objects. Unlike prior methods relying on manual prompt design, HRS-Bench constructs specific hallucination templates (\emph{e.g.}, ``Describe a scene about 3 apples and 3 dining tables'') and expands them using GPT into more complex prompts (\emph{e.g.}, ``Three apples are sitting on top of three dining tables in a cozy dining room''). Evaluation uses the AC-T2I metric, which captions the generated image and computes alignment with the original text prompt with an I2T model.\par

T2I-CompBench~\citep{huang2023t2i} is a comprehensive open-world benchmark for compositional T2I generation. It contains 6,000 text prompts across three main categories (attributes, relations, complex compositions) and six subcategories (color, shape, texture, spatial, etc.), derived from templates, existing datasets, or ChatGPT generation. Evaluation uses a multi-tool approach: BLIP-VQA assesses object attributes, UniDet evaluates spatial relationships, and a mean metric combining CLIPScore, BLIP-VQA, and UniDet scores measures complex compositions. Additionally, VLMs (\emph{e.g.}, MiniGPT-4) assess overall image-text alignment. This approach improves the precision of hallucination detection in T2I models.\par

TIFA v1.0~\citep{Hu2023TIFAAA} constructs a text prompt dataset for evaluation by sampling from the image descriptions of MSCOCO~\citep{lin2014microsoft}, containing 4K text inputs. It utilizes a language model to pre-generate corresponding question-answer pairs for each prompt, covering diverse elements. This process yields 25K questions spanning 4.5K element types. Moving beyond coarse-grained evaluation, TIFA employs VQA tools for fine-grained assessment across 11 hallucination categories (\emph{e.g.}, object type, count, attributes). The resulting VQA accuracy is reported as the TIFA Score. Experiments show that TIFA Score has a higher correlation with human judgments than CLIPScore.\par

I-HallA v1.0~\citep{lim2025evaluating} constructs a dataset of real image-text pairs sourced from scientific and historical textbooks, with a focus on factuality hallucinations. The texts are enriched with external knowledge and hallucination reasoning using GPT-4's pre-trained capabilities. An LLM then generates MCQs based on the augmented texts, while T2I models are tasked with producing corresponding images. During evaluation, a vision-language model answers MCQs associated with each image-text pair, and the I-HallA Score is calculated based on its accuracy. This MCQ-based framework offers great evaluation stability compared to caption.

T2I-FactualBench~\citep{huang2024t2i} evaluates factuality hallucinations in generated images. It covers 1,600 knowledge concepts across eight domains and defines three prompt difficulty levels: basic factual combinations (\emph{e.g.}, objects sized according to real-world facts), instantiated factual combinations (\emph{e.g.}, a table with specific foods), and instantiated knowledge concepts with interactions (\emph{e.g.}, a dog playing fetch with a hat). Evaluation uses multi-round VQA and assigns three scores: Concept Factuality, Task Completeness, and Composition Factuality, enabling granular assessment of factuality hallucinations.\par

WISE (World Knowledge-Informed Semantic Evaluation)~\citep{niu2025wise} evaluates factuality hallucinations in T2I tasks using complex prompts across natural sciences, spatiotemporal reasoning, and cultural knowledge. It contains 1,000 prompts spanning 25 subfields, sourced from educational materials, encyclopedias, common sense Q\&A datasets, and LLM-generated data. Human annotators refine the prompts for clarity, complexity, and unambiguous real-world answers. Evaluation uses the WiScore metric, a weighted sum of GPT-4o-assigned scores for consistency, realism, and aesthetic quality, providing a semantically rich assessment of image generation.\par

GenAI-Bench~\citep{li2024evaluating} is a T2I benchmark with 1,600 manually constructed text prompts, covering common hallucination types such as attributes, spatial relations, and actions. GenAI-Bench also introduces complex real-world scenarios involving counting, comparisons, and textual negation. All prompts are refined with the assistance of GPT. Evaluation combines human scoring with VQAScore metric. VQAScore quantifies image-text consistency by computing the probability of a ``Yes'' response from a VQA model to the question ``Does this figure show ‘\{text\}’?''. VQAScore exhibits higher correlation with human than CLIPScore on GenAI-Bench, validating its effectiveness for hallucination assessment.\par

\textbf{Summary of T2I Hallucination Benchmarks.} Recent evolution in T2I benchmarks highlights three key trends. First, the scope has broadened from semantic faithfulness to external factuality, evaluating adherence to both prompt constraints and world knowledge (\emph{e.g.}, T2I-FactualBench~\cite{huang2024t2i}). Second, the involvement of LLMs enables automated benchmark construction. In HRS-Bench~\citep{bakr2023hrs}, text prompts are generated based on templates first and then expanded by GPT. Third, leveraging VLMs facilitates fine-grained evaluation, exemplified by TIFA v1.0~\citep{Hu2023TIFAAA}, which employs VQA models to detect object- and attribute-level hallucinations.

Despite recent progress, several challenges remain. Although some pretrained T2I models in specialized domains have emerged~\citep{abaid2024synthesizing}, existing hallucination benchmarks still pay insufficient attention to domain-specific hallucinations. Moreover, many benchmarks still rely on coarse-grained scoring provided by VLMs~\citep{niu2025wise,li2024evaluating}. Exploring more fine-grained evaluation methods represents a promising direction for future research.

\textbf{Analysis of T2I Hallucination Causes.}
While I2T and T2I models share some common sources of hallucination, such as biases and insufficiency in training data, T2I models exhibit distinct challenges due to the generative nature of image synthesis. 
These include difficulty handling negative prompts, structural inconsistencies caused by iterative generation, and pixel-level attention ambiguities that can blend or omit objects. 
We organize the causes into two categories, emphasizing the factors most specific to T2I generation: (1) \textit{Data Limitations} and (2) \textit{Model-Specific Mechanisms}.

\begin{itemize}
\item \textit{Data Limitations.}
Mirroring I2T limitations, T2I models are affected by data sparsity and bias. 
In particular, negative prompts such as ``the bathroom area \emph{without} a curtain'' are underrepresented in common datasets~\citep{bui2025nein}, \emph{i.e.}, MSCOCO 0.43\%~\citep{lin2014microsoft}, InstructPix2Pix 0.02\%~\citep{brooks2023instructpix2pix}, CC3M 1.63\%~\citep{sharma2018conceptual}, causing T2I models to ignore or incorrectly generate negated objects~\citep{conwell2024relations}. 
Beyond negation, coverage gaps in complex compositional and factual scenarios frequently lead to generative failures~\citep{huang2023t2i,lim2025evaluating}.
Moreover, distributional biases in object co-occurence, style, or demographics frequently override prompt instructions, resulting in stereotyped or fabricated outputs~\citep{vice2025exploring}.

\end{itemize}

\begin{itemize}
\item \textit{Model Architecture and Attention.}
T2I hallucinations are closely tied to specific architectural vulnerabilities. The text encoder acts as a semantic bottleneck, and pre-trained encoders like CLIP often exhibit ``bag of words'' behavior,
leading to incorrect comprehension of complex prompts~\citep{koishigarina2025clip}. The U-Net backbone with long skip connections can propagate unrefined noise from early layers 
to the decoder, bypassing semantic guidance and producing structural inconsistencies such as disjointed limbs or incoherent textures~\citep{huang2023scalelong}. 
Meanwhile, diffusion dynamics often fail to separate semantically distant concepts, leading to mode interpolation and hybrid objects~\citep{aithal2024understanding}. 
Finally, cross-attention failures, such as attribute bleeding and token suppression, cause fine-grained object omisions and bleeding~\citep{chefer2023attend,phung2024grounded}.
These visual manifestations distinguish T2I hallucinations structurally from the textual errors observed in I2T models.

\end{itemize}

\subsection{Relationship between Hallucination Benchmarks for I2T and T2I Models}

\label{beni2tt2i}

Although hallucination evaluation originated earlier and is more comprehensive in I2T tasks, T2I benchmarks centered on image-text consistency are now gaining increasing attention. Both directions share similar trends, including the broadening of hallucination categories, increasing automation in benchmark construction, the use of model-based evaluation methods, and a stronger focus on fine-grained analysis. However, existing benchmarks capture only a subset of real-world hallucination scenarios. They are invaluable for controlled studies, but they often simplify the task (\emph{e.g.}, binary questions~\citep{Guan2023HallusionBenchAA}, limited answers~\citep{zhai2023halle}) and restrict the domain (\emph{e.g.}, common objects in single images~\citep{Li2023EvaluatingOH,chen2024multi}). Recent benchmarks are beginning to fill these gaps by considering multi-turn dialogues~\citep{cao2024visdiahalbench}, long-text~\citep{qiu2024longhalqa}, and hallucination induction~\citep{liu2025phd} in I2T and open-world generation~\cite{huang2023t2i} and real-world scenarios~\citep{li2024evaluating} in T2I. These advances aim to better capture the complexity of hallucinations encountered in real-world applications. Broadening hallucination evaluation to incorporate factual content and world knowledge remains an important direction for future work. 

In terms of construction and evaluation strategies, there is a growing overlap in the use of models across both tasks. For instance, I2T benchmarks such as VHtest~\citep{huang2024visual} and PhD~\citep{liu2025phd} employ T2I models to generate or modify images for building datasets. Conversely, T2I benchmarks like TIFA v1.0~\citep{Hu2023TIFAAA} and T2I-FactualBench~\citep{huang2024t2i} use VLMs to evaluate hallucinations. These practices suggest two complementary developments: the use of synthetic image data in I2T benchmarks, and the application of VLM-based evaluation in T2I benchmarks. They highlight the potential for mutual enhancement between I2T and T2I research in advancing hallucination evaluation.

\subsection{Hallucinations of MLLMs in Key Domains}
MLLMs show great promise, yet they face significant reliability issues in safety-critical contexts. This section analyzes the occurrence and mitigation of hallucinations across high-stakes domains, such as embodied AI, medical diagnostics, remote sensing, and agriculture.

\begin{itemize}

\item \textit{Embodied AI.}
I2T models increasingly drive manipulation, navigation, and autonomous systems~\citep{jiang2022vima,majumdar2022zson,guo2024vlm}.
Frameworks like VIMA~\citep{jiang2022vima} and Robotool~\citep{xu2023creative} generate autoregressive plans for unseen objects and creative tool use, while MUTEX~\citep{shah2023mutex} enables adaptive human-robots interaction. These capabilities extend to assistive technologies, with systems like WalkVLM~\citep{yuan2024walkvlm} and EgoBlind~\citep{xiao2025egoblind} further extend VLMs to visually impaired assistance, providing real-time navigation and egocentric understanding for the visually impaired. Despite these advancements, hallucinations constitute a critical safety bottleneck, frequently distorting scene understanding or triggering unsafe actions. This risk is particularly acute in autonomous driving, where benchmarks like NuScenes-QA~\citep{qian2024nuscenes} reveal that grounding failures often cap model accuracy below 70\%~\citep{yin2021center,jiao2023msmdfusion}. To address this, emerging mitigation strategies target both signal processing and reasoning. 
At the signal level, \cite{long2025low} apply truncated SVD to filter hallucination-prone residuals, boosting traffic captioning accuracy to 87\%. Procedurally, multi-turn reasoning frameworks like Socratic Planner~\citep{shin2024socratic} and RAT~\citep{wang2024rat} employ iterative self-correction and retrieval-augmented thought to verify plans and ensure robust execution.

\item \textit{Medical Imaging.} 
There have been a large number of medical I2T models, such as Med-Flamingo~\citep{moor2023med}, RadFM~\citep{wu2025towards}, HuatuoGPT-Vision-7b~\citep{chen2024towards}, and LLaVA-Med~\citep{li2023llava}. Despite their potential, these models are susceptible to severe hallucinations that critically undermine diagnostic trustworthiness. In clinical contexts, these errors manifest as fabricated lesions or exaggerated pathologies. Mitigation efforts prioritize specialized instruction tuning and uncertainty quantification to restore reliability. A notable advancement is MedHallTune~\citep{yan2025medhalltune}, a dataset comprising over 100k images and 1M instruction pairs designed to contrast hallucinated and non-hallucinated responses. Fine-tuning on this curated data significantly improves grounding, boosting performance on standard benchmarks like VQA-RAD~\citep{lau2018dataset} 
and SLAKE~\citep{liu2021slake}. Alongside uncertainty-aware evaluation frameworks~\citep{chen2024detecting}, these data-centric interventions are essential for ensuring the clinical reliability of medical VLMs.

\item \textit{Remote Sensing.} Remote sensing technology leverages high-altitude imagery to analyze terrestrial phenomena, facilitating applications in environmental monitoring, agriculture, and disaster management. The integration of I2T models has advanced capabilities in this domain, enabling tasks such as  image captioning~\citep{zhao2021high}, scene classification~\citep{liu2024remoteclip},
and complex reasoning~\citep{wang2024earthvqa}. However, the complexity of visual content and unique bird's perspective in remote sensing images often lead
I2T models to generate hallucinations. 
Generic models like BLIP-2~\citep{li2023blip} and MiniGPT-4~\citep{zhu2024minigpt} struggle significantly in this context, achieving less than 54\% accuracy on the RSVQA benchmark due to persistent fabrication errors~\citep{hu2025rsgpt}. To address these limitations,
the community has developed specialized fine-tuning datasets~\citep{wang2023samrs,du2019visdrone} and hallucination-focused benchmarks~\citep{li2025ddfav}. While fine-tuning on these domain-specific resources partially mitigates the issue, developing robust and reliable models for safety-critical remote sensing tasks remains an ongoing challenge.

\item \textit{Agriculture.} 
I2T models show promise in optimizing agricultural production through pests and diseases understanding~\citep{truong2025insect,li2025vllfl}, 
soil analysis~\citep{hong2024spectralgpt}, and robotic navigation~\citep{zhao2025agrivln,zhou2024navgpt}. Despite these capabilities, 
environmental distribution shifts remain a primary driver of hallucination. The high variability in crop types, weather, and farming practices 
across regions creates a domain gap that degrades zero-shot performance. For instance, models frequently fail or hallucinate during leaf segmentation 
tasks when applied to unseen environments~\citep{chawla2021quantifying}. Addressing these failures requires robust domain adaptation strategies and localized data 
collection for model training and consistent performance across diverse agricultural settings.

\end{itemize}
\textbf{Summary of MLLM Hallucinations in Key Domains.}
While MLLMs offer transformative potential in specialized domains, their deployment is impeded by critical safety risks. In high-stakes fields, model failures translate directly into real-world harms, such as physical dangers in embodied AI and misdiagnoses in medical imaging. We argue that generic mitigation strategies are insufficient for these contexts, necessitating the development of robust, domain-specific interventions~\cite{ojha2025navigating,noorani2025human}. To address this, we advocate for a sociotechnical framework that fuses expert-curated nuance, external verification, and human oversight. By prioritizing collaborative reliability over pure autonomy, this approach secures the essential trust required for deployment in safety-critical sectors.

\begin{table*}[!t]
\caption{Summary of Hallucination Detection Methods in I2T, T2I, and Unified Scenarios}
\label{detection}
\centering
\resizebox{0.93\linewidth}{!}{
\begin{tabular}{llccccc}
\hline
\multirow{3}{*}{\textbf{Task}} & \multirow{3}{*}{\makecell[c]{\textbf{Method}}} & \multirow{3}{*}{\makecell[c]{\textbf{Access Type}}} & \multicolumn{3}{c}{\makecell[c]{\textbf{Faithfulness Hallucination}\\ \textbf{Detection Granularity}}} & \multirow{3}{*}{\makecell[c]{\textbf{Detection Method}}} \\
\cmidrule(llr){4-6}
 & & & \makecell[c]{\textbf{Object-}\\ \textbf{level}} & \makecell[c]{\textbf{Scene-}\\ \textbf{level}} & \makecell[c]{\textbf{Attribute-}\\ \textbf{level}} & \\
\hline
\multirow{17}{*}{I2T}
& GAVIE~\citep{Liu2023MitigatingHI} & Black-box & \cmark & \cmark & \cmark & LLM \\
& Throne~\citep{kaul2024throne} & Black-box & \cmark & \xmark & \xmark & LLM \\
& HaELM~\citep{wang2023evaluation} & Black-box & \cmark & \cmark & \cmark & LLM \\
& FaithScore~\citep{Jing2023FaithScoreFE} & Black-box & \cmark & \cmark & \cmark & VQA \\
& \makecell[l]{CutPaste \& Find~\citep{nguyen2025cutpaste}} & Black-box & \cmark & \cmark & \cmark & VQA \\
& FactCheXcker~\citep{heiman2025factchexcker} & Black-box & \cmark & \cmark & \xmark & VQA\\
& M-HalDetect~\citep{Gunjal2023DetectingAP} & Black-box & \cmark & \cmark & \cmark & Caption \\
& AI-Feedback~\citep{xiao2025detecting} & Black-box & \cmark & \cmark & \cmark & Caption \\
& HalLocalizer~\citep{park2025halloc} & Black-box & \cmark & \cmark & \cmark & Caption \\
& CHAIR~\citep{rohrbach2018object} & Black-box & \cmark & \xmark & \xmark & Word Matching \\
& NOPE~\citep{lovenia2024negative} & Black-box & \cmark & \xmark & \xmark & Word Matching \\
& VL-Uncertainty~\citep{zhang2024vl} & Black-box & \cmark & \cmark & \cmark & Uncertainty \\
& \makecell[l]{Image2Text2Image~\citep{huang2025image2text2image}} & Black-box & \cmark & \cmark & \cmark & Similarity \\
& DHCP~\citep{zhang2024dhcp} & White-box & \cmark & \xmark & \xmark & Attention \\
& OPERA~\citep{huang2024opera} & White-box & \cmark & \xmark & \xmark & Attention\\
& HallE-Control~\citep{zhai2023halle} & White-box & \cmark & \cmark & \cmark & Feature \\
& PROJECTAWAY~\citep{jianginterpreting} & White-box & \cmark & \xmark & \xmark & Logits\\
& ContextualLens~\citep{phukan2025beyond} & White-box & \cmark & \cmark & \cmark &  Logits\\
\hline

\multirow{9}{*}{T2I}
& $\text{SR}_{\text{2D}}$ \& VISOR~\citep{gokhale2022benchmarking} & Black-box & \cmark & \cmark & \xmark & Detector \\
& DALL-Eval~\citep{Cho2022DALLEVALPT} & Black-box & \cmark & \xmark & \xmark & Detector\\
& VPEval~\citep{cho2023visual} & Black-box & \cmark & \cmark & \xmark & Detector \& VQA\\
& TIFA~\citep{Hu2023TIFAAA} & Black-box & \cmark & \cmark & \cmark & VQA\\
& T2I-CompBench~\citep{huang2023t2i} & Black-box & \cmark & \cmark & \cmark & VQA \& Detector \& Caption\\
& GraphQA~\citep{qin2024evaluating} & Black-box & \cmark & \cmark & \cmark & VQA\\
& DSG~\citep{cho2024davidsonian} & Black-box & \cmark & \cmark & \cmark & VQA \\
& VQAScore~\citep{lin2024evaluating} & Black-box & \cmark & \cmark & \cmark & VQA\\
& LLMScore~\citep{lu2023llmscore} & Black-box & \cmark & \cmark & \cmark & Detector \& Caption\\
\hline

\multirow{2}{*}{Unified}
& UNIHD~\citep{Chen2024UnifiedHD} & Black-box & \cmark & \xmark & \cmark & VQA\\
& CSN~\citep{fei2024fine} & Black-box & \cmark & \cmark & \cmark & VQA\\
\hline
\end{tabular}
}
\end{table*}

\begin{figure*}
    \centering
    \includegraphics[width=0.95\textwidth]{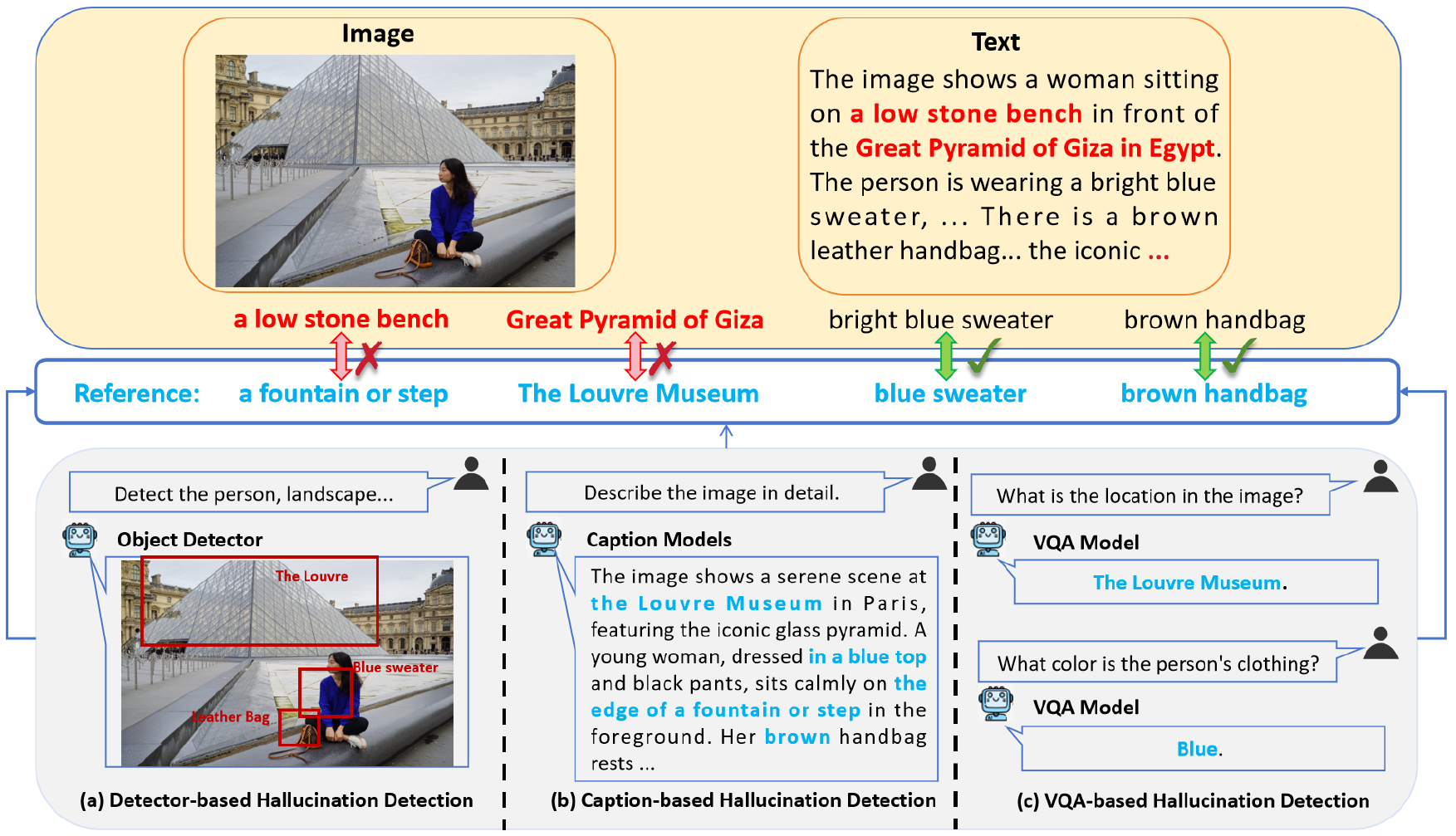}
    \caption{Examples of detection methods. At the top is a mismatched image-text pair, and hallucinated text contents are highlighted in red. At the bottom are representative (a) Detector-based, (b) Caption-based, and (c) VQA-based hallucination detection methods arranged from left to right. The reference labels are highlighted in blue.}
    \vspace{-5 pt}
    \label{detection_img}
\end{figure*}

\section{Hallucination Detection Methods}
\label{sec:det}
In this chapter, we systematically review the existing hallucination detection methods for both I2T and T2I models. Tab.~\ref{detection} summarizes the key details of the methods, including detection type, granularity, and technical approach. Fig.~\ref{detection_img} summarizes three common hallucination detection methods for I2T and T2I models, including detector-based, caption-based and VQA-based methods. The specific I2T and T2I hallucination detection methods are presented in Sec.~\ref{deti2t} and Sec.~\ref{dett2i} respectively, and unified methods in Sec.~\ref{detuni}, with analysis of their relationships in Sec.~\ref{deti2tt2i}.

\subsection{Hallucination Detection Methods for I2T Models}
\label{deti2t}
Hallucination detection is critical for assessing VLM reliability, particularly in I2T tasks like image captioning and VQA.
Depending on the accessibility of model internals, detection strategies can be broadly categorized into black-box and white-box approaches~\citep{chakraborty2025hallucination}. Given an input image, VLMs typically produce a sequence of output tokens, token-level probabilities, and internal layer representations. 
Black-box methods rely only on textual outputs, often using external evaluators or models, while white-box methods utilize internal signals such as attention weights or embeddings.
This classification clarifies trade-offs between applicability and interpretability and reflects real-world deployment constraints.
The following sections categorize existing I2T hallucination detection methods as black-box or white-box, highlighting their assumptions, mechanisms, and key applications.

\subsubsection{\textbf{Black-Box}}

\label{i2tbla}
Black-box methods lack access to VLM internals and rely on external tools for hallucination detection. They can be grouped into four types: external model-based, word matching, uncertainty-based, and similarity-based methods. External model-based methods use LLMs, VQA, or caption models to analyze VLM outputs. Word matching compares generated text with references or input content. Uncertainty-based detection identifies hallucinations via entropy or confidence scores, assuming hallucinations correlate with higher uncertainty. Similarity-based detection employs I2T models to generate reference images and compute image-image similarity.

\textbf{VQA-based Detection.}
FaithScore \citep{Jing2023FaithScoreFE} evaluates hallucinations in generative questions generated by VLMs. It first leverages GPT to identify descriptive content in the model’s response, then breaks down this content into fine-grained atomic facts. A visual entailment model (VEM) is used to verify the consistency of these atomic facts with the visual information, helping to detect hallucinations in the response. The work utilizes FaithScore as the evaluation metric. The function $\text{s}(\text{e}_\text{j}^\text{i},\text{I})$ represents the VEM, where an output of 1 indicates that the image covers the content of the atomic fact. The final hallucination score is calculated by weighting these binary scores, shown in Tab.~\ref{tab:hallucination-metrics}.
\par

CutPaste \& Find~\citep{nguyen2025cutpaste} is a training-free framework for detecting hallucinations in VLMs, leveraging a visual knowledge base and a VQA model. The knowledge base, built from Visual Genome~\citep{Krishna2016VisualGC}, contains object crops, attributes, and relationships, serving as a structured visual reference. The detection pipeline consists of three main steps: (1) extracting a scene graph from the model-generated caption; (2) locating the mentioned objects via object detection; and (3) verifying the consistency of the caption through image similarity queries and VQA-based checks. Hallucinations are identified when inconsistencies arise between the caption, detection results, and the visual knowledge base. Experiments on POPE and R-Bench show that CutPaste \& Find achieves competitive performance compared to prior methods.\par

FactCheXcker~\citep{heiman2025factchexcker} proposes an extensible and modular pipeline to detect and mitigate object and measurement hallucinations in radiology reports generated by medical VLMs. It constructs a benchmark focused on EndoTracheal Tube (ETT) placement, leveraging the MIMIC-CXR dataset and multiple report-generation models. The framework follows a query-code-update paradigm, where measurement queries are automatically generated from original reports, solved via domain-specific code modules, and incorporated back into updated reports. Evaluation metrics include precision for existence hallucination, Mean Absolute Error (MAE) for measurement hallucination, and composite scores combining both MAE and F1-score (shown in Tab.~\ref{tab:hallucination-metrics}). Experiment results demonstrate that FactCheXcker significantly reduces hallucinations and improves both measurement accuracy and placement correctness across diverse models. Without requiring retraining, it serves as a valuable tool for enhancing clinical reliability in medical imaging AI systems.

\textbf{LLM-based Detection.}
GAVIE \citep{Liu2023MitigatingHI} employs GPT to detect hallucinations in model responses. Specifically, the method inputs dense image descriptions with bounding box coordinates into GPT, then compares these descriptions to model responses for scoring. Evaluation criteria assess two dimensions: (1) Accuracy, measuring response consistency with image content; and (2) Relevance, determining whether responses directly address instructions. Hallucination scores derived from this dual-aspect analysis identify hallucinated content. Similar strategies are adopted by CC-Eval~\citep{zhai2023halle} and MMHal-Bench~\citep{sun2024aligning}, which also rely on GPT to identify or score hallucinated contents.\par

Throne~\citep{kaul2024throne} introduces an LLM-based framework for evaluating object-level hallucinations in free-form generative responses of VLMs. Throne employs public language models to analyze VLM-generated image descriptions prompted neutrally with ``Describe this image in detail.'' Through Abstractive Question Answering (AQA), it systematically queries the presence of each object class in ground-truth annotations, generating precision-recall metrics that directly quantify hallucination severity in generative tasks. Experiments show that compared to CHAIR~\citep{rohrbach2018object}, Throne achieves lower hallucination misjudgment rates.\par

HaELM~\citep{wang2023evaluation} uses a fine-tuned specialized LLM to serve as a hallucination detector for evaluating image descriptions. To construct the training corpus, it collects hallucination instances generated by VLMs and simulated via GPT models. The detector is based on LLaMA and trained using this curated dataset. Experimental results show that HaELM achieves hallucination detection performance comparable to GPT-3.5. Moreover, HaELM can be deployed locally, offering advantages in reduced computational cost.\par

\textbf{Caption-based Detection.}
M-HalDetect \citep{Gunjal2023DetectingAP} introduces a hallucination detection framework based on a fine-tuned multi-modal reward model. It constructs a training set using approximately 4,000 MSCOCO~\citep{lin2014microsoft} images paired with captions generated by InstructBLIP, annotated at the sub-sentence level to capture fine-grained hallucinations. The reward model is initialized from InstructBLIP by replacing its final embedding layer with a classification head, and is trained to predict hallucination labels at both the sentence and segment levels. To support flexible detection, the model is trained under two settings: binary (accurate vs. hallucinated) and ternary (accurate, hallucinated, and analytical). The ternary setup allows the model to distinguish factual errors from subjective analysis, reducing confusion in ambiguous cases. Experiments show that the model effectively detects hallucinations in generated captions and generalizes well to outputs from other LVLMs such as LLaVA and mPLUG-OWL.\par

AI-Feedback~\citep{xiao2025detecting} introduces a fine-grained hallucination detection framework for VLMs by constructing a detailed AI-generated feedback dataset. The model is trained to identify hallucinated content and assess its severity across various types, including object, attribute, and relationship. To build the dataset, 5000 images are sampled from Visual Genome~\citep{Krishna2016VisualGC}, and GPT-4 is leveraged to analyze the VLM-generated caption, which detects hallucinations and provides structured feedback. The detection model is trained on triplets consisting of the image, the VLM outputs, and the corresponding GPT feedback. In addition to enabling fine-grained hallucination detection, this work proposes Hallucination Severity-Aware Direct Preference Optimization (HSA-DPO), a mitigation strategy that leverages severity labels to guide preference-based fine-tuning.\par

HalLocalizer~\citep{park2025halloc} proposes a lightweight framework for token-level hallucination detection in the process of VLM generating text responses. Trained on the HalLoc dataset (150000 token-annotated samples across VQA, instruction-following, and captioning tasks), it provides probabilistic confidence scores rather than binary decisions, addressing ambiguity in real-world scenarios. The architecture offers dual input modalities: processing either the VLMs' final hidden states or generated text through a VisualBERT encoder with four linear classification heads (for object/attribute/relationship/scene hallucination types). Designed for minimal computational overhead, HalLocalizer achieves efficient inference with low latency while generating per-token hallucination probabilities. Its plug-and-play integration preserves base model efficiency, enabling seamless deployment in existing VLMs to enhance reliability without significant resource demands.\par

\textbf{Word Matching.}
CHAIR~\citep{rohrbach2018object} and NOPE~\citep{lovenia2024negative} employ word matching paradigms for hallucination detection while implementing specialized rules for distinct hallucination types. CHAIR targets object-level hallucinations through a synonym mapping system for 80 MSCOCO~\citep{lin2014microsoft} objects (\emph{e.g.}, ``player''→``person''), involving text tokenization, singularization, and compound noun disambiguation (\emph{e.g.}, distinguishing ``hot dog'' from ``dog''). Its innovation lies in fusing dual annotation sources: segmentation labels verify object presence, while human captions accommodate referential variations (\emph{e.g.}, semantic equivalence between ``dining table'' and ``coffee table''). NOPE addresses negative semantic hallucinations via a dual evaluation framework. For Negative Pronoun (NegP) tasks, it applies rule-based accuracy to detect predefined negative pronouns ($\text{A}_{\text{NegP}} = {\text{none, nothing, nobody}}$); for non-negation tasks, it leverages METEOR metrics to assess unigram matches through weighted precision, recall, and alignment scores. Although relatively coarse, these methods offer simple and effective solutions; subsequent work such as OpenCHAIR~\citep{ben2024mitigating} builds upon this foundation by employing large language models as advanced text aligners.

\textbf{Uncertainty-based Detection.} \citet{zhang2024vl} proposes VL-Uncertainty, a black-box hallucination detection method based on model uncertainty estimation. They observe that vision-language models tend to produce consistent outputs for semantically similar inputs when confident, but show variance when hallucinations occur. To measure this, it generates paraphrased queries and perturbed images (\emph{e.g.}, Gaussian blur) and computes the entropy of model predictions across them. High entropy indicates uncertainty and potential hallucination. This method avoids manual annotations and demonstrates strong performance across 10 models and 4 benchmarks, highlighting uncertainty as a meaningful hallucination indicator.

\textbf{Similarity-based Detection.} \citet{huang2025image2text2image} introduces Image2Text2Image, a novel framework for evaluating the quality of image captioning based on image-to-image consistency, thereby removing the dependence on human-annotated references. It leverages diffusion models (\emph{e.g.}, Stable Diffusion) to assess the fidelity of generated captions in a self-supervised manner. The proposed framework consists of four main components: (1) an image captioning module that generates textual descriptions from input images; (2) an image encoder that extracts visual features; (3) a diffusion model that synthesizes images from the generated captions; and (4) a similarity module that compares the original and synthesized images in feature space. Low similarity scores suggest potential deficiencies in the captions, revealing discrepancies between the original visual content and its textual representation. This approach enables automatic evaluation without reliance on human labels. \par

\subsubsection{\textbf{White-Box}}

\label{i2twhi}
Unlike black-box methods, white-box approaches leverage internal representations for hallucination detection, providing detailed insights into the model's generation process. Based on the accessed internal components, white-box detection can be categorized as attention-based, feature-based and logits-based methods. These approaches exploit distinguishable patterns between hallucinated and faithful content to detect as well as mitigate hallucinations effectively.

\textbf{Attention-based Detection.} DHCP~\citep{zhang2024dhcp} presents a white-box hallucination detection framework that exploits cross-modal attention patterns in LVLMs. Specifically, it captures the attention weights from the language model to visual tokens during the generation of the first output token, forming a structured (token × layer × head) attention tensor. Empirical analysis on the POPE-Extended dataset reveals that despite being textually identical, hallucinated and non-hallucinated responses exhibit significantly different attention distributions, which vary by hallucination type. Building on this observation, DHCP trains a lightweight two-stage MLP-based classifier (DHCP-d): the first-stage detector prioritizes high recall to flag potential hallucinations, while the second-stage refines predictions for higher precision. This cascaded design leads to a substantial performance improvement, achieving over 93\% accuracy on the POPE-Extended test set. DHCP demonstrates that internal attention dynamics provide reliable and interpretable signals for hallucination detection without requiring access to model retraining or output references.\par

OPERA~\citep{huang2024opera} detects and alleviates hallucination in VLMs by mitigating over-reliance on sparse summary tokens during decoding. Experiments show that hallucinations correlate with knowledge aggregation patterns in self-attention, where models prioritize recent summary tokens over contextual image information. Based on this observation, OPERA introduces an over-trust penalty to suppress over-aggregation in beam-search logits and a retrospection-allocation strategy to trigger token re-selection when critical patterns emerge. Experiments show that OPERA generally improves the VLMs' performances on popular benchmarks including POPE~\citep{Li2023EvaluatingOH} and MME~\citep{fu2023mme}.

\textbf{Feature-based Detection.} HallE-Control \citep{zhai2023halle} identifies that hallucinations occur when LVLM language responses contain finer-grained details than the visual module can verify, leading to unwarranted inferences. The method introduces a control module with a trainable projection layer. This layer transforms output embeddings in a way that reduces hallucinated content by modulating the model's reliance on unverifiable details. Hallucinations are further reduced by adjusting control parameters to modulate knowledge integration. This parameter-efficient approach reduces object hallucination by 44\% compared to LLaVA-7B while maintaining object coverage, enabling controllable generation of contextual versus parametric knowledge.\par

\textbf{Logits-based Detection.}
PROJECTAWAY \citep{jianginterpreting} detects and alleviates object-level hallucination in VLMs through logits lens. This technique, originally introduced in the context of language models, involves directly mapping intermediate activations to the vocabulary space using the unembedding layer, allowing for an interpretable view of token predictions at different layers. By applying this technique to VLMs, PROJECTAWAY probes each image patch to determine the presence of objects. This approach reveals that real objects exhibit significantly higher output probabilities than hallucinated ones at the token level. These probability distributions further enable patch-level spatial localization of objects. Building on this granular analysis, the method applies knowledge erasure via linear orthogonalization of image features against hallucinated object features. Results show that targeted edits to a model’s latent representations achieve up to 25.7\% hallucination reduction on MSCOCO~\citep{lin2014microsoft} while preserving performance.\par

ContextualLens~\citep{phukan2025beyond} detects and mitigates hallucinations in VLMs through token-level analysis and patch/bounding box-level grounding. Unlike the logit lens method that relies on non-contextual unembedding layer features~\citep{jianginterpreting}, ContextualLens leverages contextual token embeddings from middle VLM layers to capture multi-token concepts and spatial relationships. This enables granular hallucination detection across complex categories (attributes, actions, OCR) by computing cosine similarity between averaged answer token embeddings and image patch embeddings. For grounding, it generates precise bounding boxes via contextual embedding alignment, advancing zero-shot segmentation to grounded VQA. Evaluated on HQH benchmark~\citep{yan2024evaluating}, the method achieves significant improvements in mAP for hallucination detection while maintaining the training-free efficiency.\par

\textbf{Summary of I2T Hallucination Detection.}
As shown in Tab.~\ref{detection}, hallucination detection primarily relies on black-box methods. Most black-box methods utilize external models to assist in detection, reducing manual labeling efforts. However, the detection results are constrained by the accuracy and generalizability of the auxiliary models~\citep{li2024reference}. To further improve the detection performance, some studies have constructed hallucination detection datasets and fine-tuned existing models as detectors for evaluation~\citep{Gunjal2023DetectingAP}. However, these approaches incur high training costs, while uncertainty-based methods require no additional training but are prone to misjudging complex cases~\citep{zhang2024vl}.\par
In comparison, white-box detection methods depend on the model’s internal signals, such as attention and feature. By accessing these internal representations, white-box methods enable the tracing of hallucination sources and offer valuable insights into the model's generation process. They help reveal the flow of information and decision-making across layers, which can aid in diagnosing and mitigating specific hallucination cases. However, due to the complexity of internal mechanisms, most current white-box methods are generally limited to coarse-grained or holistic detection, struggling to distinguish fine-grained hallucination types including relationships~\citep{xiao2025detecting,song2025hallucination}.\par

Overall, while black-box methods tend to offer better generalizability and easier deployment, both approaches face significant limitations. Existing detection efforts are primarily centered on faithfulness hallucinations. The detection of factuality hallucinations, which require reasoning over external or commonsense knowledge, remains an underexplored challenge in the field.\par

\textbf{Underexplored I2T Hallucination Detection Paradigms.} 
While black-box methods are widely used for output analysis, existing white-box approaches leverage attention, feature, and logits signals~\citep{zhang2024dhcp,huang2024opera,zhai2023halle,jianginterpreting,phukan2025beyond}, which mostly focus on local tokens or single-step generation, lacking systematic monitoring of dynamic reasoning trajectories throughout the output sequence~\citep{cafagna2023interpreting}. Joint analysis of multi-modal internal signals remains limited, and there are few systematic metrics to quantify cross-modal inconsistencies. Consequently, most current detection methods are restricted to isolated indicators or specific hallucination types, leaving fine-grained detection of complex categories, semantic relations, and spatial grounding underexplored~\citep{xiao2025detecting,song2025hallucination}. Investigating these directions could enable more transparent, interpretable, and generalizable hallucination detection in I2T models.

\subsection{Hallucination Detection Methods for T2I Models}
\label{dett2i}
Early assessments of image generation primarily focus on coarse-grained evaluations such as image quality, using metrics like FID~\citep{heusel2017gans} and CLIPScore~\citep{hessel2021clipscore}. Building on these evaluations, hallucination detection in T2I models emphasizes image-text alignment at a fine-grained level, including object types, attributes, and relationships. Current hallucination detection approaches for T2I models largely rely on black-box methods, including object detection and VLMs. This section provides a comprehensive review of existing black-box methods for hallucination detection in T2I models.

\subsubsection{\textbf{Black-Box}}

\label{t2ibla}
Black-box hallucination detection methods in T2I tasks typically rely on external tools or models to verify whether the image accurately reflects the text prompt, including object detector, VQA model and caption model.\par

\textbf{Detector-based Detection.}
$\text{SR}_{\text{2D}}$ \& VISOR \citep{gokhale2022benchmarking} leverages pretrained object detectors and spatial relationship recognition algorithms to evaluate spatial consistency in T2I tasks. Objects and their spatial relationships are automatically identified in generated images. These elements are compared against the spatial layout specified in the text prompt to assess alignment with the textual description. This automated metric reveals severe spatial relationship limitations in T2I models, while demonstrating strong alignment with human judgment.\par

DALL-Eval~\citep{Cho2022DALLEVALPT} employs object detection models to quantify consistency failures in T2I models, specifically evaluating object existence, counting, and spatial relationships. The framework utilizes synthetic training data, which is derived from MSCOCO~\citep{lin2014microsoft} and procedurally generated in Unity to train a DETR-based detector, which subsequently analyzes generated images for hallucinatory content.\par

\textbf{VQA-based Detection.}
VPEval~\citep{cho2023visual} employs a programmatic evaluation framework utilizing specialized visual modules to detect hallucinations across distinct categories, including object presence, OCR accuracy, spatial relationships, and quantitative counting. This approach dynamically assembles skill-specific evaluation pipelines, where dedicated modules (\emph{e.g.}, object detection and counting systems) are invoked to verify prompt-image alignment for targeted attributes. As an interpretable visual programming paradigm, VPEval overcomes limitations of monolithic evaluation models by providing skill-optimized assessments with explanatory justifications, demonstrating significantly stronger correlation with human judgment than conventional single model-based evaluation.\par

TIFA~\citep{Hu2023TIFAAA} leverages LLMs and VQA models to assess T2I generation fidelity. For a given text prompt, question-answer pairs are automatically generated using an LLM and filtered by a question-answer model. These questions are then posed to a VQA model using the generated image. Faithfulness is evaluated by assessing consistency between the VQA answers and the ground-truth answers derived from the text prompt. This reference-free metric enables fine-grained, interpretable evaluations, demonstrates superior correlation with human judgments, and reveals significant model deficiencies in spatial relations and object composition.\par

T2I-CompBench~\citep{huang2023t2i} evaluates compositional fidelity in synthesized images across three dimensions: object attribute, spatial relationships, and scene coherence. To assess object and attribute hallucinations, binary questions derived from text prompts are evaluated using VQA models. For spatial relationship verification, object detection combined with spatial recognition algorithms is employed. Regarding scene-level coherence, MiniGPT-4-generated descriptions are scored against textual requirements. Results show limitations of T2I models to generate complex scene compositions.

GraphQA~\citep{qin2024evaluating} proposes a method for detecting hallucinations in T2I models by integrating scene graphs and VQA models. LLMs automatically generate evaluative questions from text prompts, while object detection algorithms (\emph{e.g.}, GroundedSAM) and VLMs extract visual attributes and relationships to construct image knowledge graphs. Graph Question Answering then scores response consistency between the graphs and generated images to identify hallucinations. This structured approach establishes a quantitative, interpretable framework for evaluating image-text alignment that closely correlates with human assessment standards.\par

Davidsonian Scene Graph (DSG)~\citep{cho2024davidsonian} employs a graph-based framework with VQA to detect hallucinations in generated images. Unlike GraphQA's approach~\citep{qin2024evaluating}, this methodology integrates dependency graphs to ensure comprehensive semantic coverage while preventing contradictory or redundant question generation. Drawing on linguistic formalization theories, the construction process recursively decomposes sentences into atomic propositions implemented as directed acyclic graphs. During VQA validation, negative responses to parent questions automatically terminate dependent sub-question evaluation. DSG guarantees question uniqueness and answer consistencies, thus providing adaptable semantic coverage for robust image-text faithfulness evaluation.\par

VQAScore~\citep{lin2024evaluating} leverages VQA architectures to quantify fine-grained text-visual alignment. The framework reformulates prompt-image consistency as a binary classification task, evaluating through VQA models with queries: ``Does this figure show \{Text\}?''. Alignment scores derive from the probability of affirmative responses. To enhance detection capabilities, the method incorporates CLIP-FlanT5, a bidirectional image-question encoder fine-tuned specifically for hallucination identification. This approach establishes a state-of-the-art compositional evaluation metric that overcomes lexical ambiguity limitations in conventional methods, enabling reliable assessment of complex prompts across eight benchmarks.\par

\textbf{Caption-based Detection.}
LLMScore~\citep{lu2023llmscore} integrates object detection architectures with LLMs to identify hallucinations across object-level and scene-level granularity. Object detector localizes the visual entities, which is followed by region-to-text transformers and image captioning systems that generate localized and holistic descriptions. These descriptions are evaluated against the source text prompts via LLM-based alignment assessment to detect hallucinations. Results show that the framework demonstrates stronger correlation with human judgment than conventional metrics CLIP and BLIP.\par

\textbf{Summary of T2I Hallucination Detection.}  Hallucination detection methods for T2I models mainly evaluate the consistency between generated images and textual prompts using external or fine-tuned models. Early works leverage specialized tools such as object detector~\citep{gokhale2022benchmarking}. These methods are limited in coverage and mainly focus on coarse-granularity hallucination, such as object and spatial relationship. Another common approach is to use LLMs to decompose the prompt into relevant questions, and then employ VLMs to answer these questions based on the generated image~\citep{qin2024evaluating,cho2024davidsonian}, enabling the detection of various types of hallucinations. In addition, methods concerning factuality hallucinations are relatively few. Exploring comprehensive detection methods remains an important direction.\par

\textbf{Underexplored T2I Hallucination Detection Paradigms.} Current detection metrics are predominantly semantic, not physical. A model can generate a ``car'' that is semantically recognized by CLIP, TIFA, or VPEval, but the car might lack a shadow, have inconsistent reflections, or float slightly above the ground. This blindness to real-world constraints extends to cultural nuances, historical accuracy, and other factual contexts, where semantic correctness does not guarantee fidelity to reality~\citep{borji2023qualitative}. Moreover, critical internal issues, such as ``catastrophic neglect'' in attention heads~\citep{chefer2023attend} and cross-modal misalignment during the latent denoising process~\citep{hertz2023prompt}, remain largely unexamined by current output-based detection tools. Therefore, advancing the field requires a paradigm shift towards diagnostic methods that can probe the internal generation process. Future work should focus on methods that trace alignment during denoising, audit attention maps, and incorporate physical and cultural logic directly into the evaluation framework.

\subsection{Unified Hallucination Detection Methods}
\label{detuni}

Unified hallucination detection for I2T and T2I models is emerging as a promising direction for cross-modal evaluation. Some studies integrate both tasks via image-text alignment~\citep{Chen2024UnifiedHD,fei2024fine}. Compared to task-specific methods, unified approaches capture both visual and textual content and typically require additional external models. Recent unified methods are mainly black-box, utilizing external tools and models to analyze image-text inconsistencies.

\textbf{VQA-based Detection.}
UNIHD~\citep{Chen2024UnifiedHD} introduces a unified multi-modal framework for detecting hallucinations in T2I and I2T tasks by assessing image-text consistency across four categories: object existence, attributes, scenes, and knowledge-based factuality. The methodology decomposes text prompts into atomic facts using GPT, and then generates verification questions from multiple perspectives. For visual validation, specialized tools, including object detectors, attribute classifiers (GPT-based), and external knowledge retrievers (\emph{e.g.}, Google Search), operate in parallel to gather evidentiary support. These outputs, combined with task metadata, are integrated into a VLM for hallucination assessment. This comprehensive approach establishes a novel paradigm for multi-category hallucination detection. It overcomes the limitations of task-specific methods through tool-assisted evidence validation and enables granular evaluation via the MHaluBench benchmark.

CSN (Cross-graph Siamese Network)~\citep{fei2024fine} establishes a unified hallucination detection framework for I2T and T2I tasks using structured scene graph representations. This approach detects object-existence, attribute, and relationship hallucinations through comparative analysis of dual scene graphs. Initially, a scene graph parser constructs a Visual Scene Graph (VSG) from generated images and a Textual Scene Graph (TSG) from input prompts. Subsequently, a pretrained Cross Graph Siamese Network with cross-attention mechanisms identifies semantic discrepancies between VSG and TSG representations. Ultimately, a large language model adjudicates hallucination occurrences based on detected inconsistencies. This structured framework pioneers fine-grained multi-modal hallucination detection by resolving semantic gaps in comprehension and generation processes, outperforming baselines on MHaluBench.\par

\textbf{Summary of Unified Hallucination Detection.}  
Unified hallucination detection methods aim to provide a cross-modal framework for identifying hallucinations in both I2T and T2I tasks, often leveraging question decomposition, and graph-based reasoning, assisted by external tools and models. Existing challenges include dependence on external models, modality-specific hallucination patterns, and limited efficiency. Developing robust and efficient frameworks is a key direction for future research.

\subsection{Relationship between Hallucination Detection in I2T and T2I Models}
\label{deti2tt2i}

As shown in Tab.~\ref{detection} and analysis above, hallucination detection in I2T and T2I models shares the common goal of identifying semantic inconsistencies between the input and output modalities~\citep{Chen2024UnifiedHD,fei2024fine}. Both approaches commonly depend on external models such as VQA models (\emph{e.g.}, CutPaste \& Find, TIFA)~\citep{nguyen2025cutpaste,Hu2023TIFAAA} or large language models (\emph{e.g.}, GAVIE, LLMScore)~\citep{Liu2023MitigatingHI,lu2023llmscore} to verify cross-modal consistency.\par

In addition to these shared characteristics, a difference lies in white-box methods. I2T models take advantage of techniques such as attention analysis~\citep{zhang2024dhcp} and feature~\citep{zhai2023halle}, which enable direct inspection of the generation process and help trace the origins of hallucinated content. T2I hallucination detection primarily relies on external tools, including object detectors and VQA models, and typically lacks direct access to internal generation signals. Future work could consider incorporating white-box strategies into T2I models to enhance detection effectiveness.\par

Current unified hallucination detection methods typically check the alignment between image and text content by combining multi-modal features~\citep{Chen2024UnifiedHD,fei2024fine}. These methods combine outputs from multiple external pretrained models, allowing for fine-grained identification of hallucinations. Future research could explore new mechanisms, such as fine-grained uncertainty estimation, building effective unified frameworks that better capture hallucinations in both T2I and I2T tasks.

\section{Challenges and Future Directions}
Despite growing efforts and progress in MLLM hallucination evaluation, several critical challenges remain unresolved and underexplored.
Below, we summarize these key issues to offer guidance and suggestions for future research.

\subsection{Scalable, Rigorous and Comprehensive Hallucination Evaluation}
The rapid advancement of MLLM reasoning capabilities has significantly increased the complexity of generated outputs~\citep{mu2023embodiedgpt,Shao2024VisualCA,zhao2025cot}. This evolution underscores the urgent need for more sophisticated automatic evaluation methods that transcend simple binary or surface-level assessments of hallucination~\citep{Jing2023FaithScoreFE, Jiang2024HalEvalAU, rawte2025defining, zhou2024calibrated}. However, existing hallucination benchmarks are diversifying in task format and difficulty, and their evaluations focus on model outputs in isolation, overlooking the role of training data quality, architectural choices, and pre-training objectives in shaping hallucination behavior~\citep{kang2025evaluatology}. As a result, the understanding of hallucinations remains partial, context-dependent, and limited in scope. Therefore, the evolution of evaluation is supposed to transition from static, outcome-based to dynamic, process-oriented paradigms capable of distinguishing critical factual errors and identifying root causes from a full-cycle, comprehensive perspective.

\subsection{Explainable and Reliable Hallucination Evaluation}
Hallucinations in VLMs often result from subtle misalignments between visual and textual modalities, involving complex, non-linear, and context-dependent interactions~\citep{rudin2019stop,yang2023language}. Consequently, developing explainability methods that can disentangle and trace these multi-modal reasoning processes remains a significant challenge~\citep{ben2024lvlm, yan2024evaluating}. As mentioned in Sec.~\ref{beni2t} and Sec.~\ref{bent2i}, existing hallucination evaluation benchmarks predominantly rely on coarse-grained evaluation metrics, such as accuracy, which oversimplify the problem by ignoring the nuanced spectrum of hallucination types and severities. Moreover, as benchmarks built from known datasets,it will inadvertently overlap training data, giving an optimistic view of performance~\citep{tu2025ode}. This lack of interpretability and reliability limits insight into the origins and locations of hallucinations, impeding effective diagnosis and mitigation. Advancing standardizing benchmark formats and explainable evaluation is therefore critical for enabling root cause analysis and allow more meaningful model comparisons.

\subsection{Domain-specific Factuality Hallucination Evaluation}
Evaluating factuality hallucinations in MLLMs poses a particular challenge when applied to specialized domains such as medicine~\citep{ayaz2024medvlm} and embodied intelligence~\citep{wu2025vlm,sarch2024vlm}. Unlike general-domain scenarios, these fields demand strict adherence to complex, technical, and often regulated knowledge. However, efforts to systematically evaluate hallucinations in these domain-specific contexts remain limited. As Tab.~\ref{benchmarks} shows, while benchmarks like Med-HallMark~\citep{chen2024detecting} target medical VLMs, comparable frameworks for other specialized disciplines are largely absent. Future frameworks are expected to prioritize generalization-oriented environmental settings to rigorously assess model robustness against the unpredictable nature of practical deployment and guarantee deployment readiness in safety-critical domains.

\subsection{Unified Detection for I2T and T2I Models}
Current research on I2T and T2I hallucination detection faces notable gaps, including limited investigation and fragmented evaluation standards. However, the landscape is shifting towards unified paradigms that address hallucination analysis holistically across both modalities. Pioneering approaches like UNIHD~\citep{Chen2024UnifiedHD} demonstrate the efficacy of consistency-based paradigms that generalize across I2T and T2I tasks. Modular architectures such as VPEval~\citep{cho2023visual} employ specialized submodules to localize hallucinations at a fine-grained level. These methods, however, typically depend on external tools (\emph{e.g.}, object detectors or knowledge retrievers) for content validation, which can complicate deployment and scalability. To advance the field, future research should prioritize the development of standardized, extensible frameworks that support unified detection across modalities. Establishing these foundations will be critical for improving reproducibility, fostering cross-modal insights, and enabling more robust, interpretable multimodal systems.\par

\subsection{Hallucination Evaluation in Real-World Scenarios}
While existing benchmarks such as POPE~\citep{Li2023EvaluatingOH}, MME~\citep{fu2023mme}, HallusionBench~\citep{Guan2023HallusionBenchAA}, and Med-HallMark~\citep{chen2024detecting} offer precise measurements through well-defined ground truths, they predominantly foster an ``exam-oriented'' evaluation paradigm~\citep{kalai2025language}. This reliance on curated or synthetic datasets creates a significant virtual-to-real discrepancy: models may demonstrate high proficiency in controlled settings (\emph{i.e.} overfitting dataset) yet encounter failures when facing the ambiguous, long-tail distributions inherent to real-world applications. To bridge this gap, MLLM evaluation should evolve beyond static dataset performance and draw on advancements in the LLM domain, such as Reeval's~\citep{yu2024reeval} use of adversarial perturbations and HaluEval-Wild's~\citep{zhu2024halueval} focus on real user interactions. To ensure model reliability, the paradigm must shift from evaluating static accuracy to assessing interactive robustness. This change is indispensable for establishing a continuous feedback loop that not only identifies current errors but allows for the proactive prediction of failure modes in novel, safety-critical environments.

\section{Conclusion}\label{sec13}

Recent MLLMs have demonstrated strong capabilities in multi-modal tasks, but still suffer from the hallucination problem. This survey provides a comprehensive review of hallucination issues observed in existing MLLMs, with a particular focus on both I2T and T2I paradigms. We systematically categorize hallucinations into faithfulness and factuality types and analyze current evaluation methodologies. We highlight several key trends in the field, including the emergence of automated benchmark construction, the movement towards fine-grained evaluation, and the shift towards a unified perspective that bridges evaluation approaches between I2T and T2I tasks. Furthermore, our review of hallucination detection methods highlights the growing reliance on external tools to enable black-box detection strategies. Based on these insights, we outline several promising future directions for advancing hallucination evaluation in MLLMs. We hope this survey could provide researchers in this field with a clearer understanding of the current landscape and offer valuable guidance for future exploration.

\backmatter

\section*{Data Availability Statement}

Data availability is not applicable to this paper as no new data were created or analyzed.









\begin{appendices}






\end{appendices}



{\small

}
\end{document}